\documentclass[lettersize,journal]{IEEEtran}
\usepackage{amsmath,amsfonts,amssymb,bm}
\usepackage{algorithmic}
\usepackage{algorithm}
\usepackage{array}
\usepackage[caption=false,font=normalsize,labelfont=sf,textfont=sf]{subfig}
\usepackage{textcomp}
\usepackage{stfloats}
\usepackage{url}
\usepackage{verbatim}
\usepackage{graphicx}
\usepackage{cite}
\usepackage[normalem]{ulem}
\usepackage{xcolor}

\newcommand{\kth}{$k^\text{th}$}
\usepackage[normalem]{ulem}

\usepackage{hyperref}

\usepackage{float}

\usepackage[caption=false,font=normalsize,labelfont=sf,textfont=sf]{subfig}

\usepackage{color}

\usepackage{etoolbox}
\makeatletter
\patchcmd{\@maketitle}
  {\addvspace{0.5\baselineskip}\egroup}
  {\addvspace{-1\baselineskip}\egroup}
  {}
  {}
\makeatother

\begin{document}

\title{DP-DyLoRA: Fine-Tuning Transformer-Based Models On-Device under Differentially Private Federated Learning using Dynamic Low-Rank Adaptation}

\author{Jie Xu, 	
        Karthikeyan Saravanan,
        Rogier van Dalen,
        Haaris Mehmood,
        David Tuckey,
        Mete Ozay
        % <-this % stops a space
\thanks{Manuscript received MM DD, YY; revised MM DD, YY.}
\thanks{Jie Xu, Karthikeyan Saravanan, Haaris Mehmood, David Tuckey and Mete Ozay are with Samsung R\&D Institute UK.}
\thanks{Rogier van Dalen is with AI Center-Cambridge, Samsung.}}

\markboth{Journal of \LaTeX\ Class Files,~Vol.~14, No.~8, August~2021}%
{Shell \MakeLowercase{\textit{et al.}}: A Sample Article Using IEEEtran.cls for IEEE Journals}

\IEEEpubid{0000--0000/00\$00.00~\copyright~2021 IEEE}

\maketitle

\begin{abstract}
Federated learning (FL) allows clients to collaboratively train a global model without sharing their local data with a server.
However, clients' contributions to the server can still leak sensitive information.
Differential privacy (DP) addresses such leakage by providing formal privacy guarantees, with mechanisms that add randomness to the clients' contributions.
The randomness makes it infeasible to train large transformer-based models, common in modern federated learning systems.
In this work, we empirically evaluate the practicality of fine-tuning large scale on-device transformer-based models with differential privacy in a federated learning system.
We conduct comprehensive experiments on various system properties for tasks spanning a multitude of domains: speech recognition, computer vision (CV) and natural language understanding (NLU).
Our results show that full fine-tuning under differentially private federated learning (DP-FL) generally leads to huge performance degradation which can be alleviated by reducing the dimensionality of contributions through parameter-efficient fine-tuning (PEFT).
Our benchmarks of existing DP-PEFT methods show that DP-Low-Rank Adaptation (DP-LoRA) and its variants consistently outperform other methods.
An even more promising approach, DyLoRA, which makes the low rank variable, when naively combined with FL would straightforwardly break differential privacy.
We therefore propose an adaptation method that can be combined with differential privacy and call it DP-DyLoRA.
Finally, we are able to reduce the accuracy degradation and word error rate (WER) increase due to DP to less than 2\% and 7\% respectively with 1 million clients and a stringent privacy budget of $\bm{\epsilon}=2$.
\end{abstract}


\section{Introduction}

\IEEEPARstart{T}{oday}, transformer-based models \cite{transformer} are becoming increasingly common for a wide range of applications such as natural language understanding (NLU), automatic speech recognition (ASR) and image classification \cite{mobilevit, mobile_device_transformer}. Compared to models such as recurrent neural networks (RNNs) and convolutional neural networks (CNNs), transformer-based models are known to have several advantages such as being better at handling long-range input dependencies and more efficient for training and inference due to parallel input processing \cite{transformer}. Pre-training and fine-tuning transformers is the dominant approach for building models with state-of-the-art performance \cite{bert, gpt}. These models are particularly suitable for deployment on edge devices since they can be pre-trained on massive unlabelled data at the central server without much human effort, and only a small amount of data is required per client when fine-tuning in collaboration with other clients for downstream tasks.

Federated learning (FL) \cite{federated_learning} keeps data on clients and sends only statistics about the data to a central server, to train a centrally-held model.
Though it sounds like user privacy would be improved, much information about the data is revealed through the statistics.
To address potential privacy leakage of clients' training data, further guarantees are required.

\IEEEpubidadjcol

\begin{figure}[!h]
\vspace*{-8pt}
\centering
\includegraphics[width=1.8in]{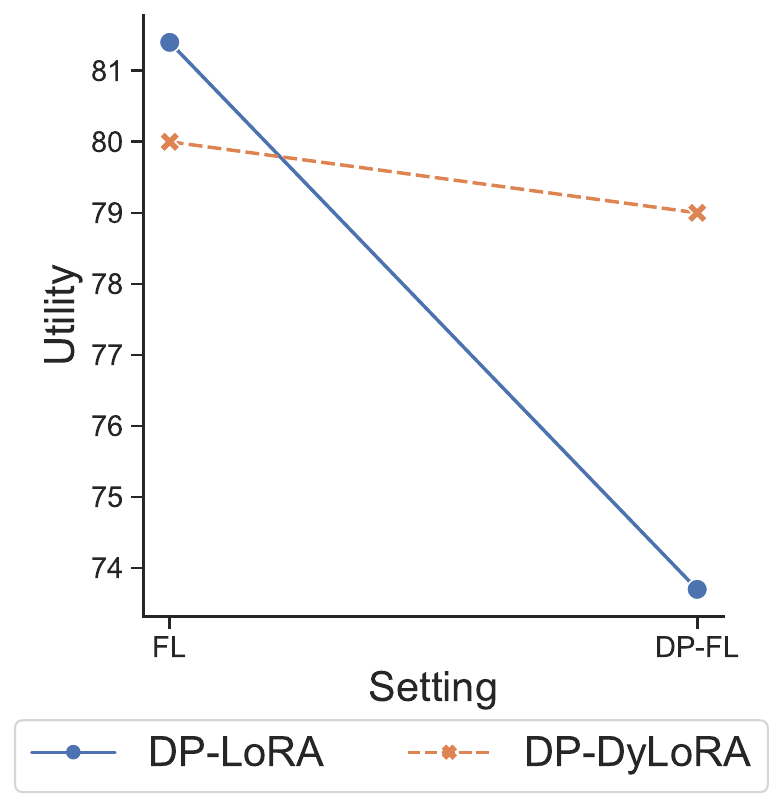}
\vspace*{-8pt}
\caption{Privacy-utility trade-offs of DP-LoRA and DP-DyLoRA on six datasets across three different domains under DP-FL. The utility is computed as the average of accuracy.}
\label{fig_lora_dylora_tradeoff}
\vspace*{-6pt}
\end{figure}

Differential privacy (DP) \cite{dp_1} is the gold standard for providing such privacy guarantees.
Very briefly, it adds so much randomness that the data gives very little away about the presence of any individual.
Naively applied to the federated learning setting, it would involve adding noise to each individual's statistics (``local DP'').
In this case, the noise would overwhelm the signal.
The alternative is to add Gaussian noise once to a sum of many contributions (``central DP'') \cite{dp-sgd, privacy_amplification_via_subsampling_1, privacy_amplification_via_subsampling_2}, and use a secure sum algorithm \cite{secure_aggregation_1, secure_aggregation_2, secure_sum} to hide individual contributions from the server.

However, this may still add too much noise.
An important lever to change this is the size of the statistics that each client sends to the server \cite{noise_magnitude_model_size_1, noise_magnitude_model_size_3}.
First, if the vector with statistics is longer, its $\ell_2$ norm will tend to be greater, and then more noise is required to hide the data.
Second, the noise needs to be added to each element of the vector, and therefore the total amount of noise increases with the length of the statistics.

To forgo the need to send a vector of the size of the model, recent works \cite{dp-fl-peft_1, dp-fl-peft_2, dp-fl-peft_3} utilise parameter-efficient fine-tuning (PEFT) methods such as Adapters and Low-Rank Adaptation (LoRA) to fine-tune transformer-based models under differentially private federated learning (DP-FL).
Only the values of a lower-rank matrix, or only the Adapter parameters, then need to be sent.
This results in much less noise being added while maintaining the same privacy guarantee, which in turn improves model performance.

Comprehensive experiments for DP-PEFT methods are missing from the literature.
Experiments in existing works \cite{dp-fl-peft_1, dp-fl-peft_2, dp-fl-peft_3} fail to address realistic system properties such as a massive number (millions) of clients in a federated learning system \cite{iot_millions_of_clients}.
Works such as \cite{dp-fl-peft_2} and \cite{dp-fl-peft_3} show experiments for only a single domain and a single type of DP-PEFT method.
\cite{dp-fl-peft_1} only considers the speech domain and evaluates Partial Embedding Updates with a combination of LoRA, without considering PEFT methods such as Adapter \cite{adapter_1, adapter_3}, Compacter \cite{compacter} and BitFit \cite{bitfit} which are often considered in works regarding PEFT and DP-PEFT methods \cite{lora, dp-lora, fl-peft_1}.

This work, on the other hand, presents a comprehensive set of experiments.
We start by empirically studying the training dynamics of fine-tuning transformer-based models via full fine-tuning on datasets of multiple domains including natural language understanding, computer vision and speech recognition.
We then show with empirical results that parameter-efficient fine-tuning can achieve much better privacy-utility trade-offs than full fine-tuning, and comprehensively benchmark existing DP-PEFT methods on three different domains under DP-FL.
The most successful PEFT scheme for DP-FL turns out to be LoRA \cite{lora,dp-lora}, which learns low-rank matrices to add to existing weight matrices.
It is fairly obvious how to use it within DP-FL, where it is called DP-LoRA.

A recent improvement, DyLoRA \cite{dylora}, proposed for NLP, does away with the manual choice of rank.
This would make it highly suitable for DP-FL, where privacy budget gets used up by hyperparameter searching.
However, a naive adaptation of DyLoRA to DP-FL straightforwardly breaks differential privacy, since users would be sending up vectors of different lengths.
In this work, we solve this conundrum by modifying the form of DyLoRA so that each cohort uses one vector length.
We show that this scheme yields the same expected value of updates as the original DyLoRA.
We call the scheme DP-DyLoRA.
Our results show that DP-DyLoRA significantly outperforms existing DP-PEFT methods including DP-Adapter \cite{adapter_1, adapter_3}, DP-Compacter \cite{compacter}, DP-BitFit \cite{bitfit}, DP-LoRA \cite{lora} and its variants DP-LoHa \cite{loha} and DP-AdaLoRA \cite{adalora} under DP-FL on datasets across three different domains.
Specifically, we show that DP-DyLoRA achieves less than 2\% accuracy drop and 7\% word error rate (WER) increase from non-private LoRA (or DyLoRA) with a strong DP guarantee ($\epsilon=2$) and 1 million clients.
DP-DyLoRA achieves noticeably better privacy-utility trade-offs than state-of-the-art DP-PEFT method DP-LoRA as shown in Figure~\ref{fig_lora_dylora_tradeoff}.

In short, the main contributions of this article are:
\begin{enumerate}
    \item First work in literature to comprehensively benchmark existing PEFT methods under DP-FL with production-level system parameters for a detailed and realistic comparison under this learning paradigm.
    \item Proposing a novel DP-FL algorithm DP-DyLoRA which achieves better privacy-utility trade-offs than state-of-the-art DP-PEFT method DP-LoRA by optimising for a range of ranks instead of a fixed rank.
    \item Formal security analysis to prove that our proposed algorithm satisfies differential privacy constraints.
\end{enumerate}

The rest of the paper is organised as follows. We first present an overview of the related work in Section~\ref{sec_related_work}, which is followed by preliminaries in Section~\ref{sec_preliminaries}. Next, we describe parameter-efficient fine-tuning with differential privacy in Section~\ref{sec_peft_with_dp} and introduce our novel DP-FL algorithm DP-DyLoRA with formal security analysis in Section~\ref{sec_dp_dylora}. We then describe our experimental setup in Section~\ref{sec_experimental_setup} and discuss our results and findings in Section~\ref{sec_experiments}. Finally, we summarise our findings in Section~\ref{sec_conclusion}.

\section{Related Work}
\label{sec_related_work}

FederatedAveraging (FedAvg) \cite{fedavg} which served as a generalisation of FederatedSGD (FedSGD) \cite{fedsgd} became a common baseline for federated learning (FL) soon after being proposed back in 2017. Following the success of FedAvg, numerous works focusing on different aspects of this learning paradigm have been published \cite{scaffold, q-ffl, fedopt}. As an important research topic for FL, various optimisation approaches were proposed to reduce the communication cost and improve robustness against non-independent and identically distributed (non-IID) data \cite{scaffold, q-ffl, fedprox, fedopt}. These methods typically attempt to tackle the communication bottleneck of FL by either reducing the number of communication rounds required for models to converge or the percentage of clients being sampled at each communication round. Works including \cite{fl_niid_1, fl_niid_2, scaffold, fedprox} focus more on model convergence with large heterogeneity in the data which is a more realistic setting for FL \cite{fl_niid_setting}.

Although federated learning allows clients to contribute to a global model without sharing local data therefore protecting data privacy to some extent, adversaries are still able to infer sensitive information from gradients sent from clients to the server \cite{dp-sgd, adversary_1, adversary_2, adversary_3}. In order to improve privacy protection in federated learning, DP-FedAvg was proposed in \cite{dp-fedavg} which adds differential privacy to the FedAvg algorithm. This is achieved by introducing noise to the uploaded gradients by using the moments accountant \cite{moments_accountant} originally proposed for differentially private stochastic gradient descent (DP-SGD) \cite{dp-sgd} with Gaussian mechanism \cite{gaussian_mechanism} and privacy amplification via subsampling \cite{privacy_amplification_via_subsampling_1}. The moments accountant provides tight privacy bounds for the sampled Gaussian mechanism \cite{moments_accountant}. There have also been recent works on training transformer models via DP-SGD which primarily focus on reducing memory and computation complexities \cite{dp-lora, ghost_clipping}.

The magnitude of the noise added to achieve differential privacy increases as the model size grows \cite{noise_magnitude_model_size_1, noise_magnitude_model_size_3}. With the current trend of developing and deploying ever larger models, parameter-efficient fine-tuning turns out to be an intuitive solution for sample-level DP-SGD as proposed in \cite{dp-lora} which potentially allows us to fine-tune less than 1\% parameters while preserving most of the model performance. Adapter \cite{adapter_1, adapter_3} and Low-Rank Adaptation (LoRA) \cite{lora} are examples of such methods which can be categorised into sequential and parallel approaches \cite{peft_sequential_parallel}, respectively. This approach has been applied also to differentially private federated learning as in \cite{dp-fl-peft_1, dp-fl-peft_2}.

\vspace*{-4pt}

\section{Federated learning with differential privacy}
\label{sec_preliminaries}

In this section, we describe federated learning with differential privacy, and explain why the number of parameters that are updated is such a crucial quantity.

\begin{table}[H]
\vspace*{-10pt}
\caption{Main nomenclature employed.}
\vspace*{-6pt}
\centering
\begin{tabular}{|c|p{6.5cm}|}
\hline
\textbf{Notation} & \textbf{Meaning}\\
\hline
$\eta$ & Learning rate\\
\hline
$\ell$ & Loss\\
\hline
$\triangledown$ & Gradient operator\\
\hline
$W^t$ & Model parameters of the global model at the start of round $t$\\
\hline
$\triangledown \ell ( W )$ & Gradients with respect to the weights $W$\\
\hline
$\Delta_k$ & Model updates of the \kth client\\
\hline
$n_k$ & Number of samples that the \kth client possesses\\
\hline
$a_{ij}$ & Number of samples that belong to class $i$ and cluster $j$\\
\hline
\end{tabular}
\vspace*{-12pt}
\end{table}

\vspace*{-8pt}

\subsection{Federated Learning}

\vspace*{-2pt}

Federated learning is a machine learning paradigm where a central server aims to train a model on data that is distributed over a large number of clients.
What the clients send is not the actual data, but instead statistics about the data.
This normally works iteratively: in each round, the server sends the most recent model to a cohort of clients.
% In Federated Averaging \cite{fedavg}, which applies to gradient-based learning, e.g.\ for neural networks, each of the clients sends to the server statistics about its data.
In Federated Averaging \cite{fedavg}, the clients trains for multiple local iterations on local data, and sends the difference between the resulting model and the original one to the server.
On the server, the average of updates from all clients turns out to form a good update for the central model, which is the key insight that allows Federated Averaging.

% multiple remote clients collaboratively train a global model on their own local data without data exchange. FedAvg \cite{fedavg} is a commonly used baseline algorithm for federated learning which has proved to be robust in many different cases \cite{fednlp}, \cite{fl_llm_size}. At the start of each round, the most updated global model is shared with clients sampled for that round. Each client then trains the model on its own local data and sends the model update back to the server. The server subsequently aggregates the client updates and applies the aggregated update to the global model without accessing client data.

The local training data in a federated learning system is not necessarily independent and identically distributed (IID) \cite{fl_niid_1}. In practice, it is very likely that individual clients train on highly skewed non-IID data \cite{fl_niid_1}. Data heterogeneity can come from different factors such as label distribution, number of samples per client and user habit.

\vspace*{-8pt}

\subsection{Differentially Private Federated Learning}

Even though in federated learning no data leaves the clients, the statistics can give away too much personal information.
The standard method for preventing this is \emph{differential privacy} \cite{dp_1, dp_2, dp_3}.
The following is a high-level introduction to differential privacy and its use in federated learning.

% \todo{epsilon, delta, Gaussian?}

Differential privacy \cite{dp_1} (DP) prevents a \emph{membership attack}, where an attacker already knows what an individual is sending, but tries to work out whether they are included in the data.
This seems like a high bar, but no meaningful lower bars have been found.
In practice, in federated learning, enough noise must be added to mask any one individual's statistics.
To do this, first the $\ell_2$ sensitivity must be constrained, which means making sure that every individual's contribution $\Delta_k$  has $\lVert \Delta_k \rVert_2 < S$, where $S$ is a scalar constant, the clipping bound.
Then, noise must be added.
The amount of noise is determined ultimately by the privacy budget $(\epsilon, \delta)$, with $\epsilon$ usually being a single-digit value, here $2$, and $\delta$ being a small fraction, here $10^{-6}$.

It would seem natural in federated learning for each client to add noise to their own data.
This is called ``local DP'' but the amount of noise would then be so high as to prevent anything from being learned.
Instead, a trick is necessary.
As originally proposed, in its ``central'' guise, DP would involve a trusted third party that would take individuals' stats in the clear, and output aggregated statistics, with noise added to each aggregate.
In the federated learning case, where the third party would merely compute a vector sum, the role of the trusted third party can be played by cryptography.

The ``Secure Aggregation'' algorithm \cite{secure_aggregation_1, secure_aggregation_2} allows many clients to contribute to a vector sum, where no one, not even the server receiving the sum, sees the individual contributions.
To guarantee central differential privacy, each client adds their part of the correct overall noise to the sum \cite{secure_sum}.
Existing secure summing algorithms are also designed to be robust to client dropouts to some extent. For simplicity, we do not consider client dropouts in this work. We also do not explicitly mention such algorithm in our experiments as results would be identical either with or without secure summing implemented. Previous works \cite{secure_aggregation_1, secure_aggregation_2, secure_aggregation_exact_sum} have shown that the server is able to receive the exact sum of client updates without access to individual updates using secure summing algorithms such as SecAgg \cite{secure_aggregation_1} and SecAgg+ \cite{secure_aggregation_2}.

Assuming such a secure summing algorithm, the differential privacy analysis first proposed in \cite{moments_accountant} can be used.
It assumed a large population of individuals, and at each round of Federated Averaging a subset of them is sampled i.i.d.\ to contribute.
The individual contributions from the selected cohort are summed with added Gaussian noise, and this is used to update the central model.
\cite{moments_accountant} proposed a DP analysis of the algorithm called the ``moments accountant'', which was much more efficient in terms of privacy budget than previous analyses, and this analysis has since been improved \cite{privacy_amplification_via_subsampling_1, privacy_amplification_via_subsampling_2}.
In the rest of this article a budget of $(2, 10^{-6})$ is used.
The Gaussian noise will be chosen to remain within this budget.

\section{Parameter-Efficient Fine-tuning with Differential Privacy}
\label{sec_peft_with_dp}

Parameter-efficient fine-tuning (PEFT) is a technique designed for efficient adaptation of pre-trained models to downstream tasks.
Instead of fine-tuning all parameters of a model, parameter-efficient fine-tuning methods aim to train only a small number of parameters.
This makes training much cheaper especially for large pre-trained models \cite{adapter_1, adapter_3, lora}.

When applied to differentially private federated learning (DP-FL), there are additional benefits to parameter-efficient fine-tuning over training the whole model.
Clients now need to send up only a smaller vector.
Less communication is then required, and the signal-to-noise ratio improves.

Here, we describe state-of-the-art PEFT methods which we include later in our benchmark. Most of these methods have been studied under DP constraints such as DP-Adapter, DP-Compacter and DP-LoRA in \cite{dp-lora}. All of the PEFT methods we describe in this section can be directly applied to the DP-FedAvg \cite{dp-fedavg} algorithm without modifying either the algorithm or the PEFT method.

\vspace*{-9pt}

\subsection{Adapter}
\label{sec_adapter}

Adapter was originally proposed in \cite{adapter_1} as an early attempt to adapter-based fine-tuning of large pre-trained models. This method reduces the number of trainable parameters by inserting a compact bottleneck adapter layer after each attention and feed-forward layer while freezing all the weights of the pre-trained model. Given a $d$-dimensional feature $x$, an adapter layer can be represented as:
\vspace*{-4pt}
\begin{equation}
\label{eqn_adapter}
\text{adapter}(x) = U(\tau (D(x))) + x,
\vspace*{-4pt}
\end{equation}
where $x \in \mathbb R^k$ is the input, $k$ is the input dimension, $U \in \mathbb R^{r \times k}$ is a linear up-projection map with rank $r$, $D \in \mathbb R^{k \times r}$ is a linear down-projection map and $\tau$ is a non-linear activation function. After the initial attempt, a few variants of Adapter have been proposed such as \cite{adapter_3} which only adds an adapter layer after the feed-forward layer. Following \cite{dp-lora}, we only consider the approach proposed in \cite{adapter_1} in our experiments.

\vspace*{-9pt}

\subsection{Compacter}
\label{sec_compacter}

Compacter proposed in \cite{compacter} introduces a more parameter-efficient version of adapter layers. This is done by replacing the dense matrices for the up-projection $U$ and down-projection $D$ with low-rank parameterised hypercomplex multiplication layer (LPHM) while removing the nonlinearity and residual connection. Each Compacter layer therefore can be represented as the sum of $n$ Kronecker products as follows:
\vspace*{-4pt}
\begin{equation}
\begin{aligned}
\label{eqn_compacter}
\text{compacter}(x) & = W_{\text{compacter}} x + b \\
  & = \bigg(\sum_{i=1}^n A_i \otimes B_i \bigg) x + b \\
  & = \bigg(\sum_{i=1}^n A_i \otimes \big(s_i t_i ^\top \big) \bigg) x + b,
% W_{\text{compacter}} =  \sum_{i=1}^n A_i \otimes B_i = (\sum_{i=1}^n A_i \otimes (s_i t_i ^\top)),
\end{aligned}
\vspace*{-4pt}
\end{equation}
where $n$ is a user-defined hyperparameter, $\otimes$ is the matrix Kronecker product, $W_{\text{compacter}} \in \mathbb R^{a \times b}$ is a Compacter layer, $A_i$ are parameters shared across all Compacter layers, $B_i$ is a low-rank matrix with non-shared parameters which is the product of two low-rank matrices $s_i \in \mathbb R^{\frac{a}{n} \times r}$ and $t_i \in \mathbb R^{r \times \frac{b}{n}}$. Here, only $B_i$ is factorised since $A_i$ are small and shared across all Compacter layers. Factorising $A_i$ therefore would degrade model performance. Since $n$ is typically set to a small value such as $n = 2$, Compacter layers therefore usually contain much fewer parameters than adapter layers.

\vspace*{-9pt}

\subsection{BitFit}
\label{sec_bitfit}

BitFit is a simple and intuitive parameter-efficient fine-tuning method where only the bias terms of the pre-trained model are fine-tuned. This method is comprehensively studied in \cite{bitfit} and is often used as a baseline method for PEFT studies \cite{compacter, lora}.

\vspace*{-9pt}

\subsection{LoRA}
\label{sec_lora}

Low-Rank Adaptation (LoRA) \cite{lora} is a parameter-efficient fine-tuning method designed for transformer-based pre-trained models. LoRA can significantly reduce the number of trainable parameters during fine-tuning by freezing the pre-trained weights and adding trainable rank decomposition matrices into each transformer layer. Let $W_\text{pt}^i \in \mathbb R^{b \times a}$ be pre-trained weight matrices of the $i^\text{th}$ layer, LoRA adds a low-rank term $B^i A^i$ with rank $r$ by:
\vspace*{-4pt}
\begin{equation}
\label{eqn_lora}
W^i_\text{LoRA} = W_\text{pt}^i + B^i A^i,
\vspace*{-4pt}
\end{equation}
where $B^i \in \mathbb R^{b \times r}$ is an up-projection and $A^i \in \mathbb R^{r \times a}$ is a down-projection. Here, $A^i$ and $B^i$ are initialised to random Gaussian noise and zero, respectively. $B^i A^i$ is therefore zero at the start of training. The pre-trained weights $W_\text{pt}^i$ are then frozen and $B^i A^i$ becomes the new trainable parameters.

LoRA has demonstrated superior performance in DP-FL both in central \cite{dp-lora} and federated learning \cite{dp-fl-peft_1} settings when compared to other parameter-efficient fine-tuning methods such as adapter \cite{adapter_1} and compacter \cite{compacter}.

\vspace*{-10pt}

\subsection{LoHa}
\label{sec_loha}

Low-Rank Hadamard Product (LoHa) or FedPara \cite{loha} is a parameter-efficient fine-tuning method proposed specifically for reducing the communication costs in federated learning. Unlike Lora, LoHa uses Hadamard product to approximate larger weight matrices as:
\vspace*{-4pt}
\begin{equation}
\label{eqn_lora}
W^i_\text{LoHa} = W_\text{pt}^i + (B_1^i A_1^i) \odot (B_2^i A_2^i),
\vspace*{-4pt}
\end{equation}
where $\odot$ denotes the Hadamard product.

\vspace*{-10pt}

\subsection{AdaLoRA}
\label{sec_adalora}

Adaptive Low-Rank Adaptation (AdaLoRA) \cite{adalora} adopts singular value pruning to adaptively optimise the rank values for different weight matrices based on the magnitude of individual singular values. The weight matrices $W^i_\text{AdaLoRA}$ for the $i^\text{th}$ layer is then defined as 
\vspace*{-4pt}
\begin{equation}
\label{eqn_lora}
W^i_\text{AdaLoRA} = W_\text{pt}^i + B^i \Lambda^i A^i,
\vspace*{-4pt}
\end{equation}
where $\Lambda^i \in \mathbb R^{r \times r}$ denotes the singular values.

\vspace*{-10pt}

\subsection{DyLoRA}
\label{sec_dylora}

A recent work of \cite{dylora} introduces a dynamic low-rank adaptation (DyLoRA), a method which aims to address two problems of the original LoRA \cite{lora}, namely, the rank of the LoRA layers are fixed after training and find an optimal rank requires an exhaustive search. This is done by training LoRA modules for a range of ranks $r \in [r_{min}, r_{max}]$ instead of a single rank. To achieve this, DyLoRA samples $b \sim p_B(.), b \in \{r_{min}, r_{min} + 1, ..., r_{max}\}$ at each training step and truncates up-projection $B$ and down-projection $A$ such that:
\begin{equation}
\begin{split}
\label{eqn_dylora}
B_b &= B[:, 1:b] \\
A_b &= A[1:b, :],
\end{split}
\vspace*{-4pt}
\end{equation}
where $B_b$ is the b-truncated up-projection and $A_b$ is the b-truncated down-projection.

\vspace*{-6pt}

\section{DP-DyLoRA}
\label{sec_dp_dylora}

\vspace*{-2pt}

%Here, we describe the proposed DP-FL algorithm DP-DyLoRA. Similar to standard DP-FL algorithms such as DP-FedAvg \cite{dp-fedavg}, DP-DyLoRA samples a portion of users at the start of each communication round before sending the latest global model to sampled users from the central server. Next, the sampled users train the model on their local data and clip the model updates to a predefined threshold before sending the clipped model updates back to the server. The clipped model updates are then aggregated, noised and applied to the global model.

%DP-DyLoRA adds new trainable LoRA modules to the model and freezes all pre-trained weights to make fine-tuning large pre-trained models more parameter-efficient. As DyLoRA, the model is also trained for a range of ranks instead of a single fixed rank. Therefore, at the start of each communication round, DP-DyLoRA samples rank $b$ uniformly at random from $\{r_{min}, r_{min} + 1, \ldots, r_{max}\}$ at the server side and shares the selected rank $b$ together with the $b$-truncated up-projection and down-projection $B_b$ and $A_b$ with the sampled users. This makes sure that all users in the same cohort update the same parameters of the global model so as to not break differential privacy. We do not consider the secondary truncation mode described in \cite{dylora} where only the $b^{\text{th}}$ rows and columns are updated since it is known to cause noticeable performance drop.

In this section, we describe the proposed algorithm DP-DyLoRA for differentially private federated learning to train LoRA weights for a variable rank. Doing so both increases signal-to-noise ratio and saves privacy budget from hyperparameter searching.
However, this runs up against a problem: the choice of rank $b$ would naturally be made per client, but this would not work with DP.
If different clients were to send up different-length vectors, this would immediately break DP.
If instead clients padded the statistics they sent up with zeros, this would decrease the signal-to-noise ratio on the highest ranks significantly.

\begin{algorithm}[!h]
\caption{DP-DyLoRA with \colorbox{pink}{SecureSum}.}
\begin{algorithmic}[1]
\STATE {\textsc{Server}}$ $
\STATE \hspace{0.5cm}\textbf{parameters}
\STATE \hspace{1.0cm}number of communication rounds $T$
\STATE \hspace{1.0cm}all users $ \mathcal{K} $
\STATE \hspace{1.0cm}user sampling rate $ q \in (0,1] $
\STATE \hspace{1.0cm}noise multiplier $ z $
\STATE \hspace{1.0cm}clip norm $ S $
\STATE \hspace{1.0cm}minimum rank $ r_{min} $
\STATE \hspace{1.0cm}maximum rank $ r_{max} $
\STATE \hspace{1.0cm}pre-trained model weights $ W^\text{0} $
\STATE \hspace{0.5cm}\textbf{for} each dense weight matrix $ W_{i}^{0} $ in $ W^\text{0} $ \textbf{do}
\STATE \hspace{1.0cm}$ B_{i}^{0} \leftarrow $ (random Gaussian initialization)
\STATE \hspace{1.0cm}$ A_{i}^{0} \leftarrow $ (zero initialization)
\STATE \hspace{1.0cm}$ W_{i}^{0} = W_{i}^{0} + B_{i}^{0} A_{i}^{0} $
\STATE \hspace{0.5cm}Freeze all pre-trained weights $ W^{0}$
\STATE \hspace{0.5cm}\textbf{for} each round $ t = 1, 2, \ldots T $ \textbf{do}
\STATE \hspace{1.0cm}Sample rank $b^t \in \{r_{min}, r_{min} + 1, \ldots, r_{max}\}$ \\\hspace{1.0cm} uniformly at random
\STATE \hspace{1.0cm}\textbf{for} each dense weight matrix $ W_{i}^{t} $ in $W^{t} $ \textbf{do}
\STATE \hspace{1.5cm}$ \Hat{W}^{t}_{i} = B_{i}^{t}[:, 1:b] A_{i}^{t}[1:b, :] $
\STATE \hspace{1.0cm}Sample a subset $\mathcal{C}^t \subseteq \mathcal{K}$ of users uniformly at\\\hspace{1.0cm} random with probability $q$
% \STATE \hspace{1.0cm}\textbf{for} each user $ k \in \mathcal{C}^t $ in parallel \textbf{do}
% \STATE \hspace{1.5cm}$ \Delta^{t}_k \leftarrow $ \textsc{UserUpdate}($ \Hat{W}^t $)
\STATE \hspace{1.0cm}$ \sigma = z \cdot S $
%\STATE \hspace{1.0cm}$ \Hat{W}^{t+1} \leftarrow \big(\Hat{W}^t + \frac1{\lvert \mathcal{C}^t \rvert}  \sum_{k \in C^{t}} \Delta^{t}_k \cdot \min\big(1, \frac{S}{\lVert \Delta^{t}_k \rVert}\big)$ \\\hspace{2.4cm} $+ \mathcal{N}(0, I\sigma^2)\big)$
\STATE \hspace{1.0cm}$ W^{t+1} \leftarrow W^t + \frac1{\lvert \mathcal{C}^t \rvert} \textsc{SecureSumDP}\big($
\STATE \hspace{1.5cm} $ \{ \textsc{UserUpdate}(k, \Hat{W}^t)\}_{k \in \mathcal C^t}, z, S\big) $
\STATE
\STATE \colorbox{pink}{{\textsc{SecureSumDP}}$(\{ \Delta_k \}_{k \in \mathcal C^{'}},z,S)$}
%\STATE \hspace{0.5cm}\COMMENT{This is implemented so the server cannot access $\Delta_k$}
\STATE \hspace{0.5cm}\textbf{parameters}
\STATE \hspace{1.0cm}$ \sigma = z \cdot S $
\STATE \hspace{0.5cm}\textbf{return} $ \mathcal{N}(0, I\sigma^2) + \sum_{k \in C^{'}} \Delta_k \cdot \min\big(1, \frac{S}{\lVert \Delta_k \rVert_2} \big)$
\STATE
\STATE {\textsc{UserUpdate}}$(\Hat{W}^{'})$
\STATE \hspace{0.5cm}\textbf{parameters}
\STATE \hspace{1.0cm}number of local epochs $E$
\STATE \hspace{1.0cm}minibatch size $\beta$
\STATE \hspace{1.0cm}learning rate $\eta$
\STATE \hspace{0.5cm}$ \Hat{W}^{+} \leftarrow \Hat{W}^{'} $
\STATE \hspace{0.5cm}\textbf{for} each local epoch $ e = 1, 2, \ldots E $ \textbf{do}
\STATE \hspace{1.0cm}$ \mathcal{B} \leftarrow $ (split local data into batches of size $\beta$)
\STATE \hspace{1.0cm}\textbf{for} batch $ b \in \mathcal{B} $ \textbf{do}
\STATE \hspace{1.5cm}$ \Hat{W}^{+} \leftarrow \Hat{W}^{+} - \eta \triangledown \ell ( \Hat{W}^{+} ) $
% \STATE \hspace{1.5cm}$ W_\text{trun(k)} = W_\text{trun(k)} - \eta \triangledown \ell ( W_\text{trun(k)} ) $
% \STATE \hspace{1.5cm}$ W_\text{trun(k)} = W_\text{trun}^{'} + $  FlatClip($ W_\text{trun(k)} - W_\text{trun}^{'}, S $)
\STATE \hspace{0.5cm}\textbf{return} $ \Hat{W}^{+} - \Hat{W}^{'}$
\end{algorithmic}
\label{alg_dp_dylora}
\end{algorithm}

\begin{figure}[!h]
\centering
\includegraphics[width=3.5in]{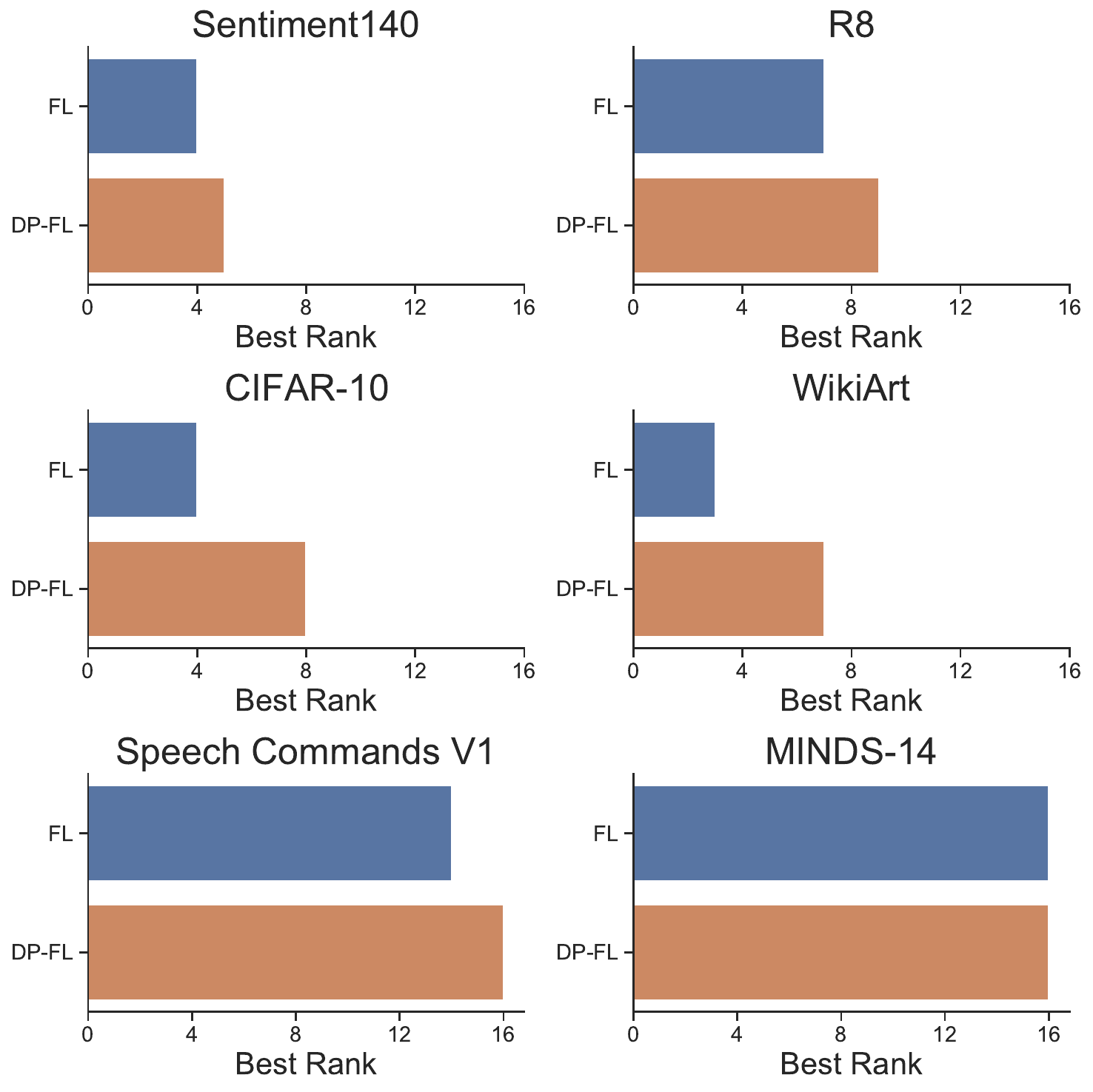}
\vspace*{-18pt}
\caption{The optimal rank values of DP-DyLoRA for the last communication round as opposed to that of DyLoRA under non-private federated learning.}
\label{fig_dylora_best_r}
\vspace*{-10pt}
\end{figure}

Instead, we propose that the server draws one $b^t$ per round $t$ for the whole cohort, and all devices train $B_{b^t}$ and $A_{b^t}$ as in \eqref{eqn_dylora}.
We do not consider the secondary truncation mode described in \cite{dylora} where only the $b^{\text{th}}$ rows and columns are updated since it is known to cause noticeable performance drop.
Note that the expectation of the change in parameters $B$ and $A$ in one round is the same whether rank $b$ is sampled separately on each client or once on the server.

The complete DP-DyLoRA algorithm we propose is in Algorithm \ref{alg_dp_dylora}.
Similar to standard DP-FL algorithms such as DP-FedAvg \cite{dp-fedavg}, DP-DyLoRA samples a portion of users at the start of each communication round before sending the latest global model to sampled users from the central server. Next, the sampled users train the model, here $B$s and $A$s, on their local data and clip the model updates to a predefined threshold before sending the clipped model updates back to the server. The clipped updates are then aggregated, noised and applied to the global model.

DP-DyLoRA freezes all pre-trained weights and adds new trainable LoRA modules to make fine-tuning large pre-trained models more parameter-efficient.
At client side, users train on their local data with a modified forward pass:
\begin{equation}
\label{eqn_dp_dylora}
h = W_\text{pt} x + \Delta W x = W_\text{pt} x + B_b A_b x.
\end{equation}
Here, $W_\text{pt}$ is frozen and only $B_b$ and $A_b$ are trainable. The outputs of $W_\text{pt} x$ and $B_b A_b x$ are summed coordinate-wise and are given the same input for the forward pass. The updates to the trainable parameters $B_b$ and $A_b$ are then clipped and sent back to the server for aggregation and noise addition as in DP-FedAvg. The communication cost apart from the initial transfer of $W_\text{pt}$ is therefore equivalent to DP-LoRA with $r = \frac{r_{min} + r_{max}}{2}$ on average which is approximately half of that of DP-LoRA with $r=r_{max}$ assuming that $r_{min} = 1$. Since the magnitude of the added noise grows with the number of parameters updated \cite{noise_magnitude_model_size_1, noise_magnitude_model_size_3}, DP-DyLoRA also achieves higher signal-to-noise ratio than DP-LoRA under the same DP-FL setting. Meanwhile, the same level of model expressiveness is preserved as the model architecture and number of trainable parameters remain the same for the global model.

At each communication round, only the $b$-truncated up-projection $B_b$ and down-projection $A_b$ are updated. This means that parameters of lower ranks are updated more often than those of higher ranks. For example, since $b$ is sampled uniformly at random from $\{r_{min}, r_{min} + 1, \ldots, r_{max}\}$, parameters of $r_{min}$ are always updated. As we can see from Figure~\ref{fig_dylora_best_r}, the best rank values tend to increase when differential privacy is applied. On the six chosen datasets, the best rank values under non-private FL are smaller on five and equal on one compared to DP-FL. This aligns with results from \cite{dp-lora} for DP-SGD in which the optimal rank for DP-LoRA ($r=16$) is higher than that of non-private LoRA ($r=4$).

\vspace*{-4pt}

\subsection{Security Analysis}
\label{sec_security_analysis}

Here, we analyse the security of our proposed method DP-DyLoRA and show that it satisfies $(\epsilon, \delta)$-DP.

At each communication round $t$, each sampled user in $C^t \subseteq \mathcal{K}$ sends model updates to the server after local training. Although we randomly sample the rank $b^t \in \{r_{min}, r_{min} + 1, \ldots, r_{max}\}$ to be trained at round $t$, each user sampled at the same communication round trains and shares parameters of the same rank $b^t$. Therefore, we use moments accountant \cite{moments_accountant} with R\'enyi Differential Privacy (RDP) \cite{rdp} to privatise the shared model updates for a tight composition bound. For any $\alpha \in (1,\infty)$ and $\epsilon > 0$, a randomised mechanism $\mathcal{M}$ satisfies $(\alpha, \epsilon^{\prime})$-RDP if for all neighbouring datasets $D$ and $D^{\prime}$, we have
\vspace*{-4pt}
\begin{equation}
\label{eqn_rdp}
D_{\alpha}(\mathcal{M}(D) \vert \vert \mathcal{M}(D^{\prime})) \triangleq \frac{1}{\alpha - 1} \log \mathbb{E} \left( \frac{D(x)}{D^{\prime}(x)} \right)^{\alpha} \leq \epsilon^{\prime}.
\end{equation}

A randomised mechanism $\mathcal{M}$ that satisfies $(\alpha, \epsilon^{\prime})$-RDP also satisfies $(\epsilon, \delta)$-DP with
\begin{equation}
\label{eqn_rdp_dp}
\epsilon = \epsilon^{\prime} + \log(\frac{\alpha - 1}{\alpha}) - \frac{\log \alpha + \log \delta}{\alpha - 1},
\end{equation}
for any $0 < \delta < 1$ \cite{rdp_accountant}.

Since we only consider training of large transformer-based models in this work, we use Gaussian mechanism \cite{gaussian_mechanism} instead of Laplace mechanism \cite{laplace_mechanism} for achieving $(\epsilon, \delta)$-DP. This is because Gaussian mechanism allows the use of L2 sensitivity and the L2 sensitivity of a long vector is much smaller than its L1 sensitivity, leading to much less noise being added. To satisfy $(\epsilon, \delta)$-DP with Gaussian mechanism, noise of $\mathcal{N}(0, I\sigma^2)$ is added to model updates with
\begin{equation}
\vspace*{-2pt}
\label{eqn_gaussian_mechanism}
\sigma = z S,
\vspace*{-2pt}
\end{equation}
where $\sigma$ is used to compute Gaussian noise, $z$ is the noise multiplier calculated by the moments accountant for privacy parameters such as privacy budget and sampling rate and $S$ is the clipping threshold.

We further combine the Gaussian mechanism with privacy amplification via sampling \cite{privacy_amplification_via_subsampling_1, privacy_amplification_via_subsampling_2} to reduce the added Gaussian noise $\sigma$ to
\begin{equation}
\label{eqn_privacy_amp_via_sampling}
\sigma = \frac{z S}{|C|},
\end{equation}
where $|C|$ is the cohort size.

Tens of thousands of clients need to be sampled at each communication round for training large transformer-based models under DP-FL such as the ones we use for our experiments \cite{flair}. This number is reasonable in large-scale federated learning systems \cite{iot_millions_of_clients} but is difficult to achieve in simulation using current federated learning frameworks due to the amount of computation required \cite{flair}. Following \cite{dp-fedavg, flair}, we simulate the noise level of a larger cohort size $C_{large}$ with a smaller cohort size $C_{small}$ by
\begin{equation}
\label{eqn_virtual_cohort_size}
\sigma^{+} = \frac{C_{small}}{C_{large}}\sigma^{\prime},
\end{equation}
where $\sigma^{+}$ is used to compute noise for larger cohort size and $\sigma^{\prime}$ is for smaller cohort size.

\vspace*{-4pt}

\section{Experimental Setup}
\label{sec_experimental_setup}

In this section, we present a comprehensive description of our experimental setup. This includes the details of the datasets and models used in our experiments as well as baseline and novel methods implemented.

\vspace*{-8pt}

\subsection{Datasets and Tasks}
\label{sec_datasets_and_tasks}

We set up our experiments to ensure that our results will be applicable to a wide range of domains and tasks. As shown in Table~\ref{tab:datasets}, six different datasets are used in our experiments covering various tasks in Artificial Intelligence (AI) domains including computer vision, natural language understanding and speech, which are briefly described below:

\begin{table*}[!t]
\caption{Details of the datasets used in our experiments\label{tab:datasets}.}
\vspace*{-6pt}
\centering
\begin{tabular}{|c|c|c|c|c|}
\hline
\textbf{Dataset} & \textbf{Task} & \textbf{Total Num. Clients} & \textbf{Sampled Num. Clients} & \textbf{Num. Rounds}\\
\hline
Sentiment140 & Text classification & 21876 & 100 & 300\\
\hline
R8 & Text classification & 1000 & 100 & 200\\
\hline
CIFAR-10 & Image classification & 1000 & 100 & 100\\
\hline
WikiArt & Image classification & 1000 & 100 & 200\\
\hline
Speech Commands V1 & Keyword spotting & 1503 & 100 & 300\\
\hline
MINDS-14 & Automatic speech recognition & 100 & 10 & 2000\\
\hline
\end{tabular}
\vspace*{-8pt}
\end{table*}

\begin{itemize}
\item{\textbf{Natural Language Understanding: }} Sentiment140 (sent140) is used for sentiment analysis. It consists of 1.6 million tweets from over 660,000 users. Following \cite{sent140_preprocessing}, we remove users with less than 10 samples each, leaving us with 21,876 users.

R8 is another text classification dataset which is a subset of the Reuters-21578 dataset of news articles \cite{reuters-21578} with 8 classes and over 7,000 samples.

\item{\textbf{Computer Vision: }} CIFAR-10 \cite{cifar-10} and WikiArt \cite{wikiart} are used for the task of image classification. CIFAR-10 contains 60,000 32x32 images in 10 classes, each of which with 10\% of the images. It is a labeled subset of the 80 million tiny images dataset \cite{tiny_images}.

WikiArt \cite{wikiart} consists of over 81,000 images of artworks taken from \url{WikiArt.org}. Each artwork is labeled by its artist, genre and style. We use the artist label only and remove images belonging to artists with less than 100 artworks in the dataset, leaving us with 23 artists and hence 23 classes.

\item{\textbf{Speech Recognition: }}

Speech Commands V1 (SC V1) is a keyword spotting dataset with over 64,000 audio samples produced by 1,503 different speakers. We consider each of the available labels as a different class, therefore making it a 30-class classification task.

MINDS-14 is an automatic speech recognition dataset consisting of over 1,800 audio recordings in English. Each sample in the MINDS-14 dataset is also labeled by its intent with a total of 14 intent classes.
\end{itemize}

The metric we use for MINDS-14 is word error rate (WER) which is defined as the ratio of errors made in a transcript to the total number of words. More specifically, it is computed as follows:
\begin{equation}
\text{WER} = \frac{S + D + I}{N},
\end{equation}
where S, D, and I denote the number of substitutions, deletions and insertions, respectively, and N represents the total number of words.

For all other datasets, we use accuracy as the performance measurement which is defined as:
\begin{equation}
\text{Accuracy} = \frac{TP + TN}{TP + TN + FP + FN},
\end{equation}
where TP, TN, FP, FN denote the number of true positives, true negatives, false positives and false negatives, respectively.

\vspace*{-6pt}

\subsection{Models}
\label{sec_models}

\vspace*{-2pt}

We use transformer-based pre-trained models of similar numbers of parameters (over 20 million) for our experiments as shown in Table~\ref{tab:models}. Memory consumption and speed are calculated using a single NVIDIA A10, a batch size of 1 and a maximum duration of 1 second for DistilHuBERT. In this work, we consider large models to be around 25 million parameters for deployment on edge devices as in \cite{fl_llm_size} due to memory limitations of such devices. Transformer models of similar sizes are used in works including \cite{fl_transformer_3} for non-private federated learning and \cite{fl_llm_size} for differentially private federated learning, and are suitable for deployment on mobile devices \cite{mobile_device_transformer}. Smaller models such as convolutional neural networks (CNNs) and recurrent neural networks (RNNs) are used in previous works including \cite{dp-fedavg, dp-scaffold, fl-dp-ftrl}. These models are however less capable than larger transformer-based pre-trained models and deliver sub-optimal performance for a wide range of tasks \cite{gpt, bert, vit}.

The same model is used for tasks of the same domain. Therefore, we use BERT-small \cite{bert} for experiments on Sentiment140 and R8, ViT-small \cite{vit} for CIFAR-10 and WikiArt, and DistilHuBERT \cite{hubert, distilhubert} for Speech Commands and MINDS-14. Despite the fact that these models are extremely small compared to state-of-the-art large language models with rapidly increasing sizes such as LLaMA \cite{llama, llama_2} with 70 billion parameters and GPT-4 \cite{gpt-4} with 1.7 trillion parameters, models with over 20 million parameters are considered large for either on-device deployment or differentially private federated learning.

\begin{table*}[!t]
\caption{Datasets and models used for our experiments.}
\vspace*{-6pt}
\label{tab:models}
\centering
\begin{tabular}{|l|l|c|c|c|}
\hline
\textbf{Model} & \textbf{Datasets} & \textbf{Num. Parameters} & \textbf{Memory (Training)} & \textbf{Time (Training)}\\
\hline
Bert-small & Sentiment140 \& R8 & 28.7M & 2.1GB & 0.02s\\
\hline
ViT-small & CIFAR-10 \& WikiArt & 22.1M & 2.1GB & 0.04s\\
\hline
DistilHuBERT & Speech Commands \& MINDS-14 & 23.5M & 2.3GB & 0.03s\\
\hline
\end{tabular}
\vspace*{-16pt}
\end{table*}

\vspace*{-6pt}

\subsection{Federated Learning}
\label{sec_fl_setup}

\vspace*{-2pt}

For all our experiments, we consider a centralised and cross-device federated learning setting with a central server coordinating the training process and a subset of the clients being sampled at each communication round. We do not address resolving client drift caused by data heterogeneity, client dropouts, or continual learning in which client data is not necessarily stationary.

We developed our method in PyTorch. We simulate non-private and differentially private federated learning setups on 2 NVIDIA A100s or 8 NVIDIA A10s.

\vspace*{-8pt}

\subsection{Non-IID Partitioning}
\label{sec_niid_partitioning}

\vspace*{-2pt}

We use non-independent and identically distributed (non-IID) data partitioning for all our experiments unless stated otherwise. As we can see from Table \ref{tab:datasets}, Sentiment140 and Speech Commands V1 are naturally non-IID. These datasets provide user ID for each sample which allows us to assign data produced by each unique person (e.g., speaker) to each client. It is also possible to set up the WikiArt dataset in a similar fashion by using artist as the prediction target. However, this would leave us with only 23 clients in total, making the experiments unrealistically small-scale.

We therefore choose to utilise Dirichlet distribution to artificially achieve non-IID label distribution for the remaining four datasets, including R8, CIFAR-10, WikiArt and MINDS-14. For these datasets, we partition by drawing from a Dirichlet distribution with $\alpha = 0.1$ by default following \cite{dirichlet_alpha_1, dirichlet_alpha_2, dirichlet_alpha_3} which results in each client holding samples from very few classes.

\vspace*{-8pt}

\subsection{Noise Mechanism}
\label{sec_dp_mechanism}

\vspace*{-2pt}

For model training with user-level differential privacy, we only consider the Gaussian mechanism \cite{gaussian_mechanism}. We exclude the Laplace mechanism \cite{laplace_mechanism} from our experiments because it relies on the $L1$ sensitivity. That is, the magnitude of the client updates has to be computed using the $L1$ norm of the vector. On the other hand, the Gaussian mechanism allows the use of either the $L1$ or $L2$ sensitivity. For both mechanisms the standard deviation of the added noise grows linearly with the sensitivity \cite{laplace_reason_why_exclude}. Since we use large models of around 25 million parameters in our experiments, the $L1$ norm of the model update would be extremely large which would make it impossible for the model to learn effectively.

This is true even if we apply parameter-efficient fine-tuning. For example, assuming that the model has 25 million parameters and we reduce the number of trainable parameters to $1\%$ by applying parameter-efficient fine-tuning. The model update from client $k$ will then be a vector $\Delta_k$ of size 250 thousand. For simplicity, let's also assume that $\Delta_k$ is a vector of all $0.01$. The $L1$ norm is then $\lVert \Delta_k \rVert_1 = 2500$ which is much bigger than the $L2$ norm $\lVert \Delta_k \rVert_2 = 5$. We therefore do not consider the Laplace mechanism in our experiments.

\vspace*{-8pt}

\subsection{Large Cohort Noise-Level Simulation}
\label{sec_noise_level_simulation}

\vspace*{-2pt}

Following \cite{dp-fedavg} and \cite{flair}, we simulate the noise-level of larger cohort sizes with smaller ones. In practice, differentially private federated learning (DP-FL) is applied to systems with millions of clients \cite{iot_millions_of_clients}. However, it is infeasible to simulate this many clients due to resource constraints. Since a larger cohort size $C$ leads to less noise added for the same privacy guarantee, we simulate realistically large cohort size $C_{\text{large}}$ with smaller cohort size $C_{\text{small}}$. This makes our results more meaningful in terms of practical deployment of DP-FL. We follow the approach in \cite{dp-fedavg} and \cite{flair} for achieving this. The noise $\sigma$ we use for simulation is then computed as $\sigma^{+} = \frac{C_{\text{small}}}{C_{\text{large}}} z_{\text{large}} \cdot S$ where $z_{\text{large}}$ is the noise multiplier calculated based on $C_{\text{large}}$.

\vspace*{-8pt}

\section{Experiments}
\label{sec_experiments}

\vspace*{-2pt}

We experimentally investigate full fine-tuning, existing parameter-efficient fine-tuning (PEFT) methods and our proposed method DP-DyLoRA under differentially private federated learning (DP-FL) unless otherwise stated. Our experiments cover three different domains, namely, natural language understanding, computer vision and speech in order to ensure that our results are applicable to a wide range of tasks and a variety of scenarios. We additionally study the impact of data heterogeneity on DP-FL with full fine-tuning to investigate the root causes of the significant performance drop in such learning paradigm.

\vspace*{-10pt}

\subsection{Full Fine-tuning}
\label{sec_experiments_fft}

\vspace*{-2pt}

DP-FL tends to degrade model performance due to the noise added to model updates. Previous works have proven that there is a proportional relationship between the number of updated parameters and the magnitude of the added noise \cite{noise_magnitude_model_size_1, noise_magnitude_model_size_3}. Hence, it is particularly challenging to train large models with differential privacy. Together with the potential non-independent and identically distributed (non-IID) data distribution challenge, large transformer models are likely to fail when learning under DP-FL.

The lower signal-to-noise ratio can be remedied by sampling more clients at each communication round \cite{dp-fedavg}. Therefore, assuming a constant subsampling rate of 1\%, we study the relationship between total number of clients and performance drop from FL to DP-FL for models such as BERT-small which is relatively large for deployment on edge devices \cite{fl_llm_size}. Similar participation rates have been used in works including \cite{sampling_rate_1} and \cite{sampling_rate_2}. In contrast, previous works on on-device DP-FL \cite{dp-fedavg, dp-scaffold, fl-dp-ftrl} utilise models such as recurrent neural networks (RNNs) and convolutional neural networks (CNNs) which are much smaller in size than our chosen models.

\begin{figure}[!h]
\centering
\includegraphics[width=3.3in]{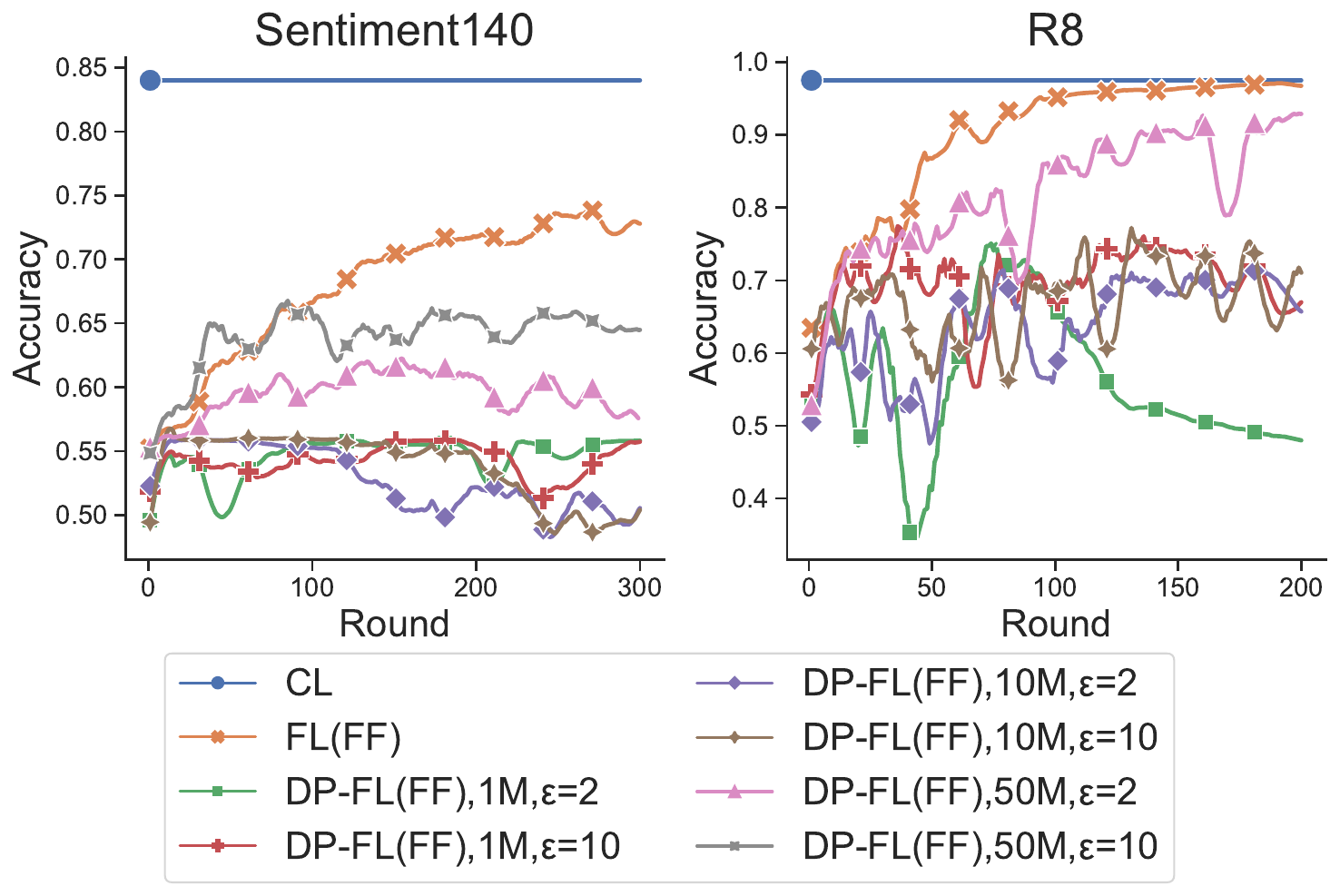}
\includegraphics[width=3.3in]{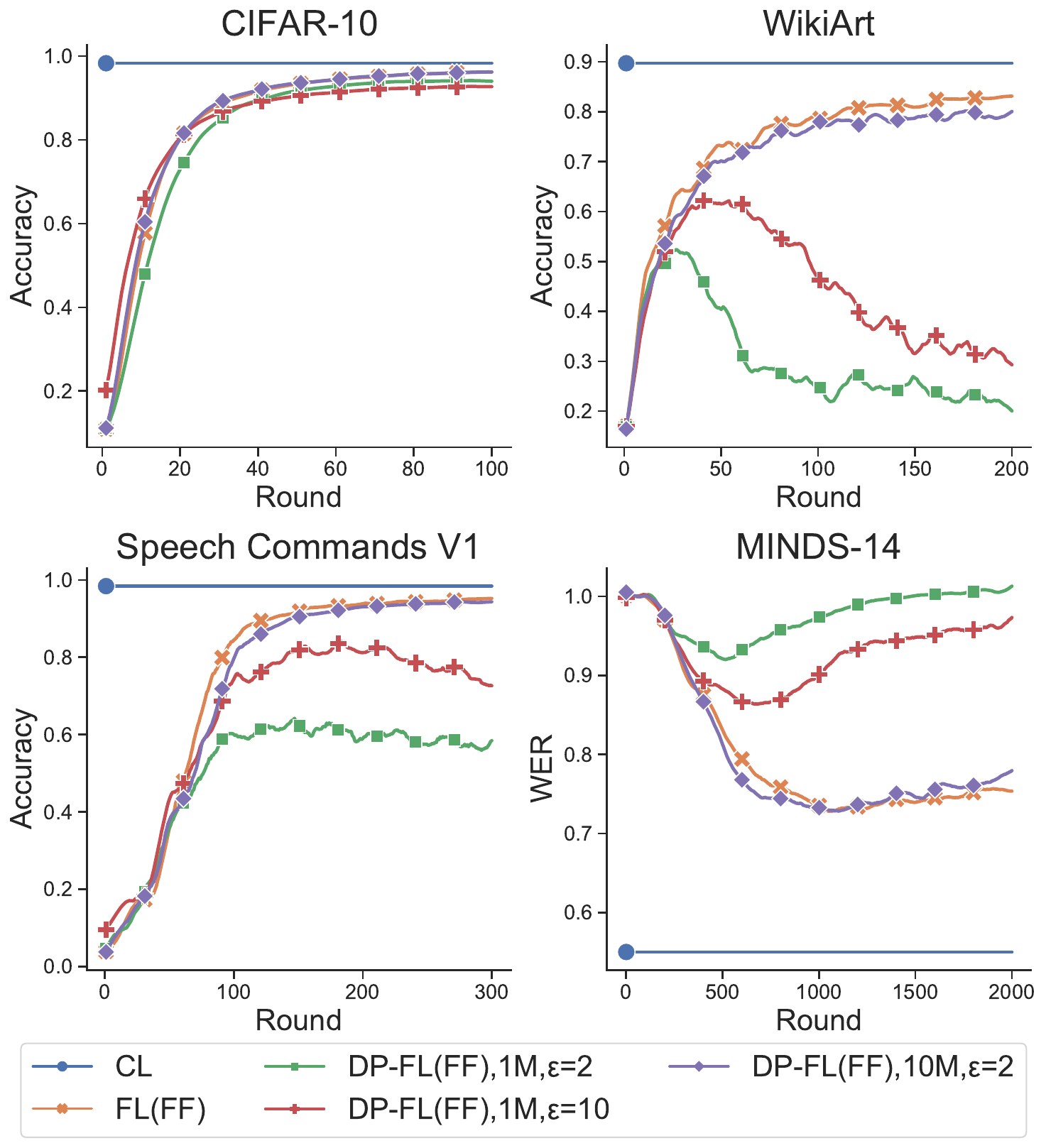}
\vspace*{-8pt}
\caption{Model performance with different number of clients in production and privacy budgets. All datasets are produced using non-IID partitioning and $\alpha=0.1$ for Dirichlet distribution if applicable. CL, FL, DP-FL denote central learning, federated learning and differentially private federated learning, respectively.}
\label{fig_q1_num_clients_privacy_budget}
\vspace*{-12pt}
\end{figure}

From Figure~\ref{fig_q1_num_clients_privacy_budget}, we can see that different tasks require utterly different settings with regard to differential privacy in order to achieve most performance of non-private federated learning. The results indicate that Sentiment140 is the most challenging task here under privacy constraints since signal seems to be completely dominated by added noise even with 50 million clients and $\epsilon=2$. The model only starts to learn after increasing the privacy budget to $\epsilon=10$. On the R8 dataset, the result is relatively close to that of non-private federated learning after increasing the number of clients to 50 million with a stringent privacy budget of $\epsilon=2$.

For the image classification task, the model achieves nearly identical result to that of non-private federated learning with only 1 million clients and $\epsilon=2$ on the CIFAR-10 dataset. This indicates that CIFAR-10 is the least challenging task amongst all we have chosen in DP-FL setting. On the other hand, the DP-FL result on WikiArt is fairly poor with 1 million clients even with a more generous privacy budget of $\epsilon=10$, and the model only produces reasonable results after increasing the number of clients to 10 million. The two datasets in the speech domain, namely, Speech Commands and MINDS-14 show similar behaviour to that of WikiArt with poor initial performance and results close to those of non-private federated learning after increasing the number of clients to 10 million.

\noindent{\bf Take-aways}

\begin{list}{}{}
\item{When training large models on-device using full fine-tuning under DP-FL, tens of millions of clients may be required for the model to learn effectively.}
\end{list}

\vspace*{-8pt}

\subsection{Data Heterogeneity}
\label{sec_experiments_data_heterogeneity}

\vspace*{-2pt}

In real-world scenarios, users possess different characteristics such as voice, interest and habit. These differences results in a highly non-independent and identically distributed (non-IID) distribution of client data in almost all cases. It is therefore essential to study the impact of data heterogeneity on model training in DP-FL.

\begin{figure}[!h]
\vspace*{-8pt}
\centering
\includegraphics[width=3.3in]{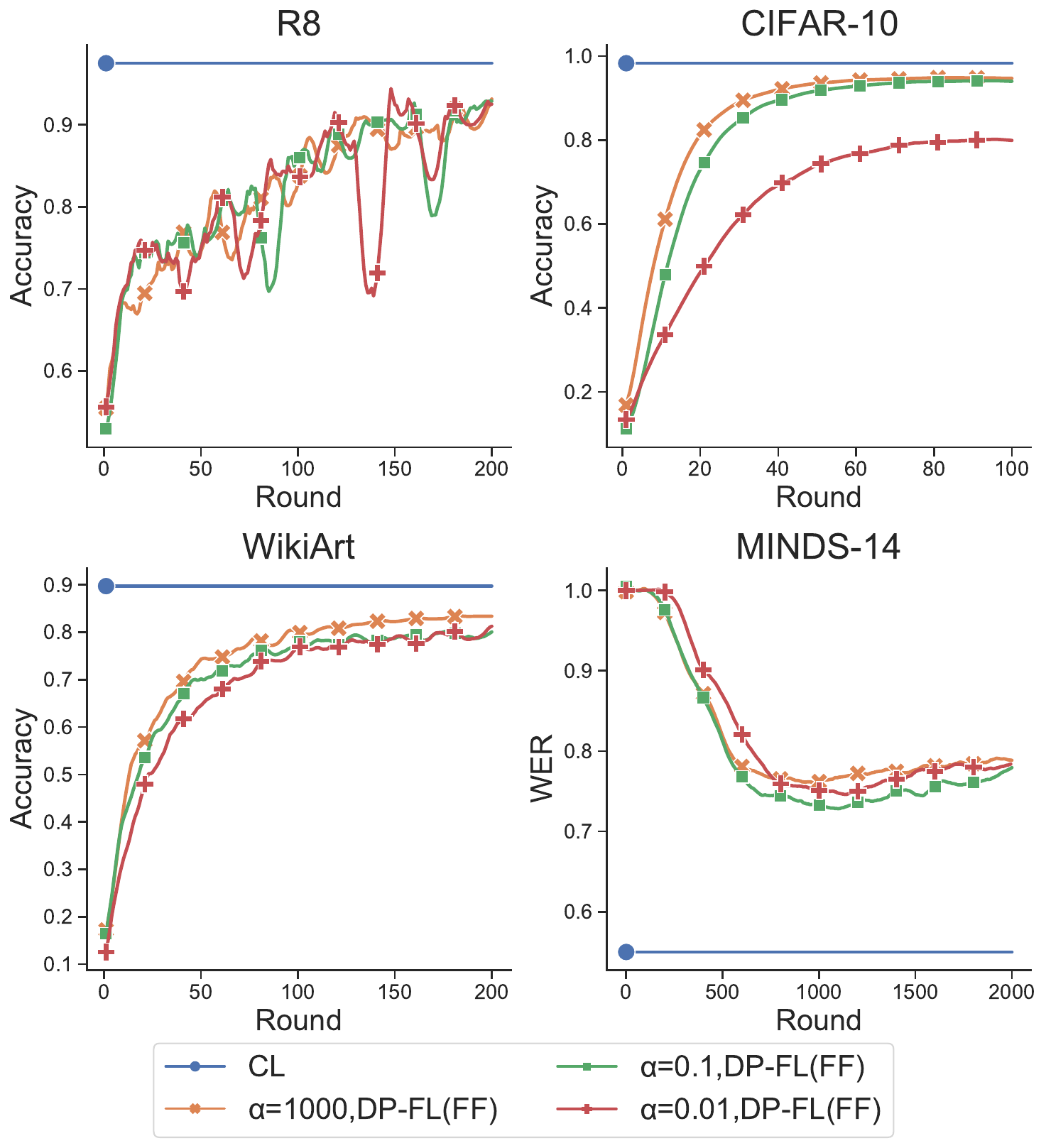}
\vspace*{-10pt}
\caption{Model performance with IID and non-IID data partitioning with the level of data heterogeneity being controlled by sampling from Dirichlet distribution.}
\label{fig_q2_dirichlet}
\vspace*{-4pt}
\end{figure}

\begin{figure}[!h]
\centering
\includegraphics[width=3.3in]{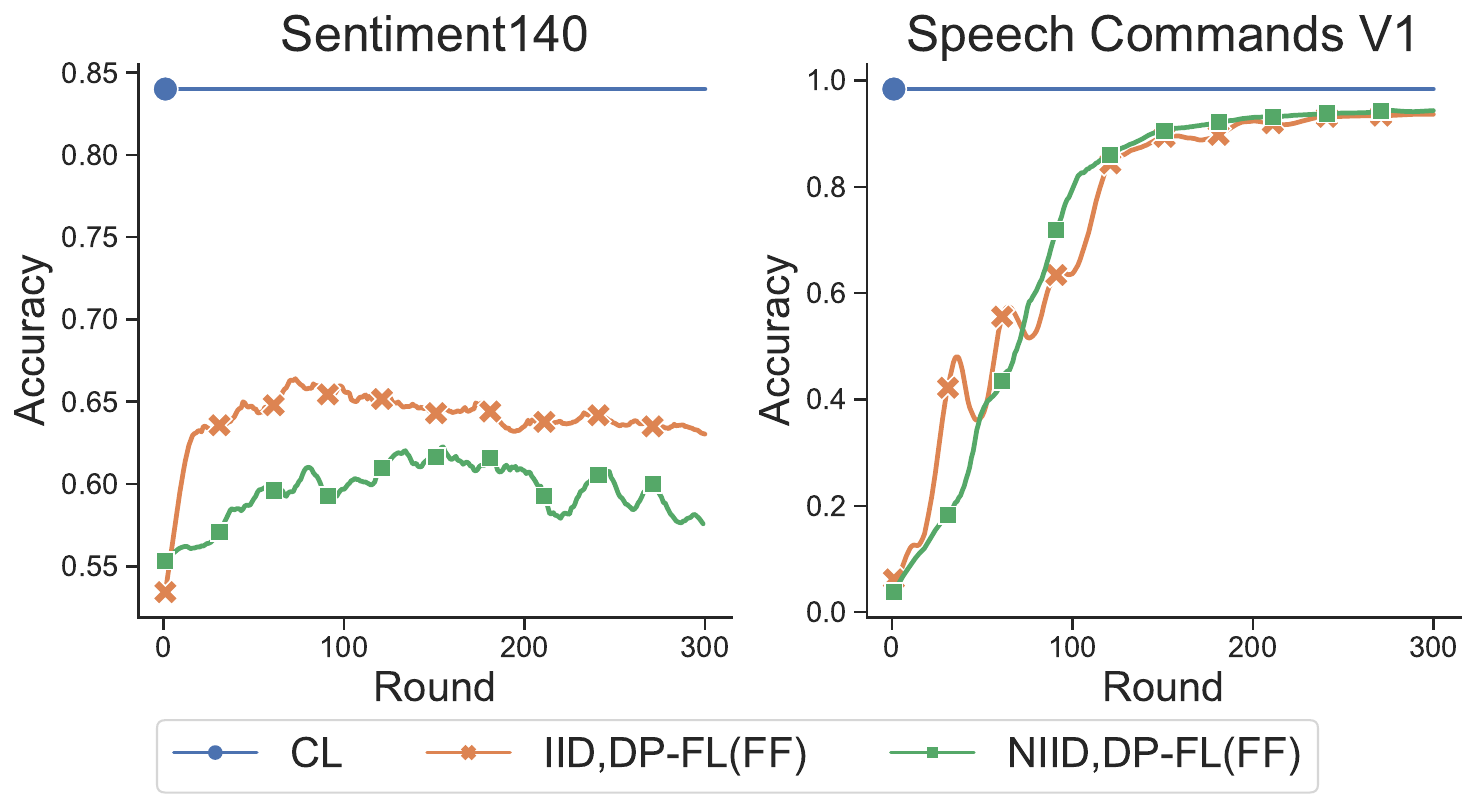}
\vspace*{-8pt}
\caption{Model performance with IID and non-IID data partitioning with the level of data heterogeneity being controlled by natural factors.}
\label{fig_q2_natural}
\vspace*{-12pt}
\end{figure}

As shown in Figure~\ref{fig_q2_dirichlet} and \ref{fig_q2_natural}, we train the model on each task with both IID and non-IID data partitioning using a single combination of client number and privacy budget taken from Q1. In Figure~\ref{fig_q2_dirichlet}, we show results with $\alpha=[0.01,0.1,1000]$ for R8, CIFAR-10, WikiArt and MINDS-14 following \cite{dirichlet_alpha_1, dirichlet_alpha_2, dirichlet_alpha_3} which are non-IID by drawing labels from a Dirichlet distribution. When $\alpha$ is set to $0.1$, there is hardly any difference from IID distribution with $\alpha=1000$ except for WikiArt on which the accuracy drops by approximately 3\%. After further increasing the level of data heterogeneity by decreasing $\alpha$ to $0.01$, results are roughly the same on R8 and WikiArt. However, on CIFAR-10 and MINDS-14, model performance degrades by approximately 14\% in accuracy and 10\% in word error rate, respectively. This is caused by severe client drift due to data heterogeneity.

For Sentiment140 and Speech Commands which are both non-IID by natural factors, we can see from Figure~\ref{fig_q2_natural} that the model achieves similar performance on the latter with both IID and non-IID distributions. This is likely due to the similar sample distributions by class. However, on the Sentiment140 dataset, the model performs noticeably better under IID data partitioning with approximately 4\% improvement. Since Sentiment140 is a binary classification dataset, some clients may only hold samples of a single class even if it is not partitioned to have a non-IID label distribution. This leads to a relatively high level of data heterogeneity which is realistic in practice due to users having different interests and habits.

These results indicate that on most datasets such as R8, CIFAR-10 and Speech Commands, there is no noticeable gap between model performance with IID and non-IID data distribution under DP-FL assuming a reasonable level of data heterogeneity. When working with extremely skewed non-IID data where each client possesses samples of a single class only, a significant performance drop can sometimes be observed such as in the case of CIFAR-10 and MINDS-14. Other than this special case, our results show that the performance drop in DP-FL is mainly caused by the added noise for DP guarantee rather than data heterogeneity.

\noindent{\bf Take-aways}

\begin{list}{}{}
\item{Data heterogeneity may further degrade model performance under DP-FL. This leads to worse privacy-utility trade-offs.}
\end{list}

\vspace*{-8pt}

\subsection{Parameter-efficient Fine-tuning}
\label{sec_experiments_peft}

\vspace*{-2pt}

Recent works \cite{dp-lora, dp-fl-peft_1, dp-fl-peft_2} have started utilising parameter-efficient fine-tuning (PEFT) methods to fine-tune transformer models with differential privacy via both central and federate learning. Apart from the obvious benefits of lower computation and communication cost, the primary motivation originates from the proven fact that fewer trainable parameters lead to better privacy-utility trade-offs \cite{noise_magnitude_model_size_1, noise_magnitude_model_size_3}.

We thereby start with comparing model training via full fine-tuning and PEFT with the same number of clients and privacy budget. Here, we use LoRA as an example of PEFT methods as it's empirically proven to be superior to other popular PEFT methods on natural language understanding tasks in \cite{dp-lora} and is used in \cite{dp-fl-peft_1} for research on DP-FL.

\begin{figure}[!h]
\centering
\includegraphics[width=3.3in]{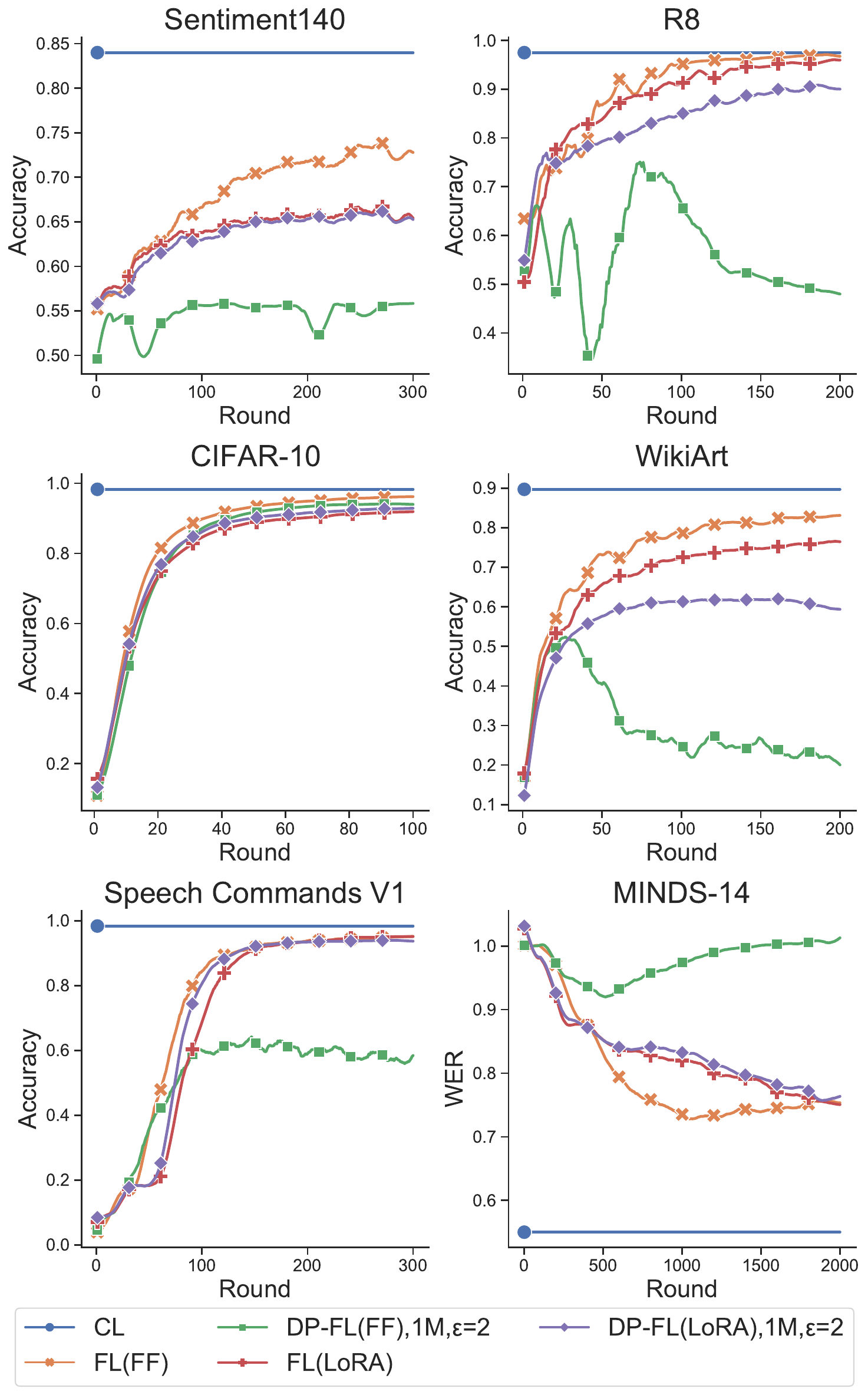}
\vspace*{-8pt}
\caption{Model performance with different number of clients in production and privacy budgets.}
\label{fig_q3_num_clients_privacy_budget}
\vspace*{-12pt}
\end{figure}

As shown in Figure~\ref{fig_q3_num_clients_privacy_budget}, private models fine-tuned with LoRA significantly outperform those obtained via full fine-tuning on all tasks except for CIFAR-10, where the performance gap between non-private and differentially private training is relatively small. However, on other tasks such as Sentiment140, R8 and Speech Commands, parameter-efficient fine-tuning unlike full fine-tuning allows us to achieve most performance of non-private federated learning with only 1 million clients and a stringent privacy budget of $\epsilon=2$. Moreover, when LoRA is applied, the number of trainable parameters decreases to approximately 1\% of that of full fine-tuning. This not only helps reduce computation cost on both the server and client devices but also makes communication between the server and clients much cheaper. Since only the trainable parameters need to be shared, this also serves as an effective solution for potential communication bottleneck when deploying large models on edge devices.

Next, we benchmark exising PEFT methods under FL and DP-FL. We experiment with PEFT methods including Adapter, Compacter, BitFit and LoRA which are often considered in existing works on parameter-efficient fine-tuning \cite{lora, dp-lora, fl-peft_1}. We also include results for two popular variants of LoRA, namely LoHa \cite{loha} and AdaLoRA \cite{adalora} to make our benchmark more comprehensive. Our benchmark covers three different domains including natural language understanding (NLU), computer vision (CV) and speech, similar to our previous experiments. Regarding datasets, we again choose to use Sentiment140/R8 for NLU, CIFAR-10/WikiArt for CV and Speech Commands V1/MINDS-14 for the speech domain.

\textbf{Hyperparameter:} For federated learning and privacy parameters, we use 1 million users, a subsampling rate of 1\%, $\epsilon$=2, and $\delta$=1e-6 for all six datasets. For clipping threshold, we search over three values \{0.1, 1.0, 10.0\}. For Sentiment140, we train for 300 rounds and search over five learning rates \{5e-2, 1e-1, 2e-1, 5e-1, 1e-0\}. For R8, we train for 200 rounds and search over four learning rates \{1e-1, 2e-1, 5e-1, 1e-0\}. For CIFAR-10, we train for 100 rounds and search over four learning rates \{2e-2, 5e-2, 1e-1, 2e-1\}. For WikiArt, we train for 200 rounds and search over four learning rates \{5e-2, 1e-1, 2e-1, 5e-1\}. For Speech Commands V1, we train for 300 rounds and search over four learning rates \{2e-1, 5e-1, 1e-0, 2e-0\}. For MINDS-14, we train for 2000 rounds and search over four learning rates \{2e-2, 5e-2, 1e-1, 2e-1\}. Regarding parameters for DP-Adapter, DP-Compacter, DP-LoRA, DP-LoHa and DP-AdaLoRA, we choose to use $r$=16 for all methods and additionally $n$=8 for DP-Compacter and half of the initial rank as target rank for DP-AdaLoRA which are derived from previous works including \cite{lora, dp-lora, adalora}.

\begin{table*}[!t]
\caption{Accuracy of parameter-efficient fine-tuning methods on five classification datasets.}
\vspace*{-6pt}
\label{tab_peft_results_classification}
\centering
\begin{tabular}{|l|l|l|l|l|l|l|l|l|l|}
\hline
\multicolumn{1}{|c|}{\textbf{Method}} & \textbf{Sent140} & \textbf{R8} & \textbf{Trained params} & \textbf{CIFAR-10} & \textbf{WikiArt} & \textbf{Trained params} & \textbf{SC V1} & \textbf{Trained params} & \textbf{Avg.}\\
\hline
Adapter (w/o DP) & 70.9 & 95.4 & 0.93\% & 94.5 & 67.6 & 2.06\% & 90.0 & 2.09\% & 83.6\\
DP-Adapter & 70.7 & 86.0 & 0.93\% & 45.2 & 47.7 & 2.06\% & 87.9 & 2.09\% & 67.5\\
Compacter (w/o DP) & 65.5 & 77.2 & 0.059\% & 36.3 & 48.6 & 0.15\% & 89.5 & 0.90\% & 63.4\\
DP-Compacter & 65.3 & 76.2 & 0.059\% & 35.1 & 43.6 & 0.15\% & 86.9 & 0.90\% & 61.4\\
BitFit (w/o DP) & 72.5 & 94.7 & 0.096\% & 96.5 & 77.4 & 0.25\% & 94.6 & 0.94\% & 87.1\\
DP-BitFit & 66.3 & 77.6 & 0.096\% & 89.9 & 49.7 & 0.25\% & 59.7 & 0.94\% & 68.6\\
\hline
LoRA (r=16, w/o DP) & 73.6 & 96.0 & 0.48\% & 95.3 & 81.0 & 1.37\% & 95.5 & 2.11\% & 88.2\\
DP-LoRA (r=16) & 70.4 & 90.0 & 0.48\% & 90.6 & 61.7 & 1.37\% & 92.5 & 2.11\% & 81.0\\
LoRA (r=8, w/o DP) & 73.3 & 97.0 & 0.25\% & 94.9 & 81.1 & 0.71\% & 96.3 & 1.91\% & 88.1\\
DP-LoRA (r=8) & 70.7 & 92.0 & 0.25\% & 92.6 & 63.5 & 0.71\% & 92.9 & 1.91\% & 82.3\\
LoRA (r=1, w/o DP) & 73.0 & 81.4 & 0.056\% & 95.2 & 82.2 & 0.12\% & 96.1 & 1.73\% & 85.5\\
DP-LoRA (r=1) & \textbf{72.5} & 80.3 & 0.056\% & 90.3 & 58.8 & 0.12\% & 85.1 & 1.73\% & 77.4\\
\hline
LoHa (r=16, w/o DP) & 72.4 & 92.6 & 0.90\% & 96.2 & 77.3 & 2.66\% & 92.8 & 1.66\% & 86.2\\
DP-LoHa (r=16) & 70.0 & 91.5 & 0.90\% & 94.5 & 60.5 & 2.66\% & 91.7 & 1.66\% & 81.6\\
AdaLoRA (r=16, w/o DP) & 71.0 & 91.8 & 0.48\% & 96.4 & 77.2 & 1.37\% & 93.6 & 2.11\% & 86.0\\
DP-AdaLoRA (r=16) & 70.5 & 90.0 & 0.48\% & \textbf{95.0} & 58.9 & 1.37\% & 92.3 & 2.11\% & 81.3\\
\hline
DyLoRA (w/o DP) & 72.0 & 96.6 & 0.48\% & 94.8 & 77.4 & 1.37\% & 95.5 & 2.11\% & 87.2\\
DP-DyLoRA & 72.0 & \textbf{96.2} & 0.48\% & 94.4 & \textbf{75.5} & 1.37\% & \textbf{93.9} & 2.11\% & \textbf{86.4}\\
\hline
\end{tabular}
\vspace*{-12pt}
\end{table*}

\begin{table}[!t]
\caption{WER of parameter-efficient fine-tuning methods on the MINDS-14 dataset for automatic speech recognition.}
\vspace*{-6pt}
\label{tab_peft_results_asr}
\centering
\begin{tabular}{|l|l|l|}
\hline
\multicolumn{1}{|c|}{\textbf{Method}} & \textbf{MINDS-14} & \textbf{Trained params}\\
\hline
Adapter (w/o DP) & 68.4 & 1.35\% \\
DP-Adapter & 68.7 & 1.35\% \\
Compacter (w/o DP) & 64.5 & 0.14\% \\
DP-Compacter & 67.6 & 0.14\% \\
BitFit (w/o DP) & 76.1 & 0.18\% \\
DP-BitFit & 85.2 & 0.18\% \\
\hline
LoRA (r=16, w/o DP) & 51.7 & 0.62\% \\
DP-LoRA (r=16) & 69.3 & 0.62\% \\
LoRA (r=8, w/o DP) & 53.1 & 0.41\% \\
DP-LoRA (r=8) & 75.9 & 0.41\% \\
LoRA (r=1, w/o DP) & 80.8 & 0.23\% \\
DP-LoRA (r=1) & 81.6 & 0.23\% \\
\hline
LoHa (r=16, w/o DP) & 56.2 & 1.66\% \\
DP-LoHa (r=16) & 59.7 & 1.66\% \\
AdaLoRA (r=16, w/o DP) & 56.9 & 2.11\% \\
DP-AdaLoRA (r=16) & 58.9 & 2.11\% \\
\hline
DyLoRA (w/o DP) & 55.6 & 0.62\% \\
DP-DyLoRA & \textbf{58.0} & 0.62\% \\
\hline
\end{tabular}
\vspace*{-12pt}
\end{table}

\textbf{Results:} Our benchmarking results covering DP-Adapter, DP-Compacter, DP-BitFit, DP-LoRA, DP-LoHa and DP-AdaLoRA across three different domains are shown in Table~\ref{tab_peft_results_classification} and \ref{tab_peft_results_asr}. On Sentiment140, DP-Adapter achieves the best accuracy of 70.7\% which is marginally higher than 70.4\% and 70.5\% achieved by DP-LoRA and DP-AdaLoRA, respectively. On R8, DP-LoHa gives the best accuracy of 91.5\%, followed by 90.0\% for both DP-LoRA and DP-AdaLoRA. For image classification on CIFAR-10 and WikiArt, DP-AdaLoRA achieves the best accuracy of 95.0\% for the former and DP-LoRA for the latter with an accuracy of 61.7\%. On CIFAR-10, both DP-Adapter and DP-Compacter suffer from slow and unstable convergence which subsequently leads to much worse accuracy achieved. On Speech Commands V1, DP-LoRA once again achieves the best accuracy of 92.5\%. The word error rates (WERs) of 58.9\% and 59.7\% achieved by DP-AdaLoRA and DP-LoHa respectively are better than the rest.

% Since our benchmarking results from Section~\ref{sec_experiments_peft} show that DP-LoRA and its variants achieve significantly better performance under DP-FL than other DP-PEFT methods, we run additionally experiments for LoRA under non-private federated learning to investigate the performance drop caused by providing DP guarantee. As we can see from Table~\ref{tab_peft_results_classification} and \ref{tab_peft_results_asr}, the average accuracy of 81.0\% achieved by DP-LoRA is noticeably lower than the average accuracy of 88.2\% under non-private FL.

\noindent{\bf Take-aways}

\begin{list}{}{}
\item{Overall, DP-LoRA and its variants outperform other existing DP-PEFT methods under DP-FL for training large transformer-based models on-device. However, noticeable performance degradation can still be observed for DP-LoRA with a strong privacy budget of $\bm{\epsilon}$=2 and 1 million clients (over 7\% in accuracy and over 17\% in WER).}
\end{list}

\vspace*{-8pt}

\subsection{DP-DyLoRA}
\label{sec_experiments_dylora}

We therefore propose DP-DyLoRA, which has better privacy-utility trade-offs than DP-LoRA under DP-FL due to fewer trainable parameters to be shared in most communication rounds.

Like DyLoRA \cite{dylora}, DP-DyLoRA trains LoRA weights for a variable rank instead of a fixed rank. When applied to DP-FL, each sampled client will train the LoRA weights for the same rank that is within a predefined range at each communication round. This means that all clients sampled at the same round will update exactly the same parameters of the model in order to provide DP guarantee. We empirically prove that DP-DyLoRA outperforms existing DP-PEFT methods including DP-LoRA under DP-FL.

\textbf{Hyperparameter:} We opt for the same federated learning and privacy parameters, and search over the same clipping thresholds and learning rates as in Section~\ref{sec_experiments_peft}. As for parameters specific to DyLoRA, we set minimum and maximum ranks to $r_{min}$=1 and $r_{max}$=16. For DyLoRA, we perform evaluation at the server side at the end of every 10 rounds since each rank between $r_{min}$ and $r_{max}$ needs to be evaluated. On Sentiment140 and R8, we apply gradient clipping to DyLoRA under non-private FL as well since the model fails to converge otherwise. We additionally experiment with $r$=\{1, 8\} with $\epsilon$=2 for both DP-LoRA and DP-DyLoRA for a more comprehensive comparison.

\textbf{Results:} As we can see from Table~\ref{tab_peft_results_classification} and \ref{tab_peft_results_asr}, DP-DyLoRA achieves an accuracy of 72.0\% on Sentiment140 which is noticeably better than all existing DP-PEFT methods except for DP-LoRA with $r$=1 which achieves a marginally better accuracy of 72.5\%. On CIFAR-10, the accuracy of 94.4\% achieved by DP-DyLoRA is marginally lower than those of DP-LoHa and DP-AdaLoRA. On the other datasets including R8, WikiArt, Speech Commands V1 and MINDS-14, DP-DyLoRA outperforms all other DP-PEFT methods including DP-LoRA with $r$=\{1, 8, 16\}. Overall, DP-DyLoRA achieves a much better performance under DP-FL with an average accuracy of 86.4\% as opposed to the 77.4\%, 82.2\% and 81.0\% average accuracy achieved by DP-LoRA with $r$=\{1, 8, 16\}, respectively. Similarly, for automatic speech recognition (ASR) on MINDS-14, the WERs of 81.6\%, 53.1\% and 51.7\% achieved by DP-LoRA with $r$=\{1, 8, 16\} respectively are significantly outperformed by DP-DyLoRA with a WER of 58.0\%. We highlight the best accuracy or WER under DP-FL for each task and the best average accuracy across all five tasks in Table~\ref{tab_peft_results_classification} and \ref{tab_peft_results_asr}.

\begin{figure*}[!t]
    \centering
    \subfloat[\label{a}]{%
        \includegraphics[width=0.39\linewidth]{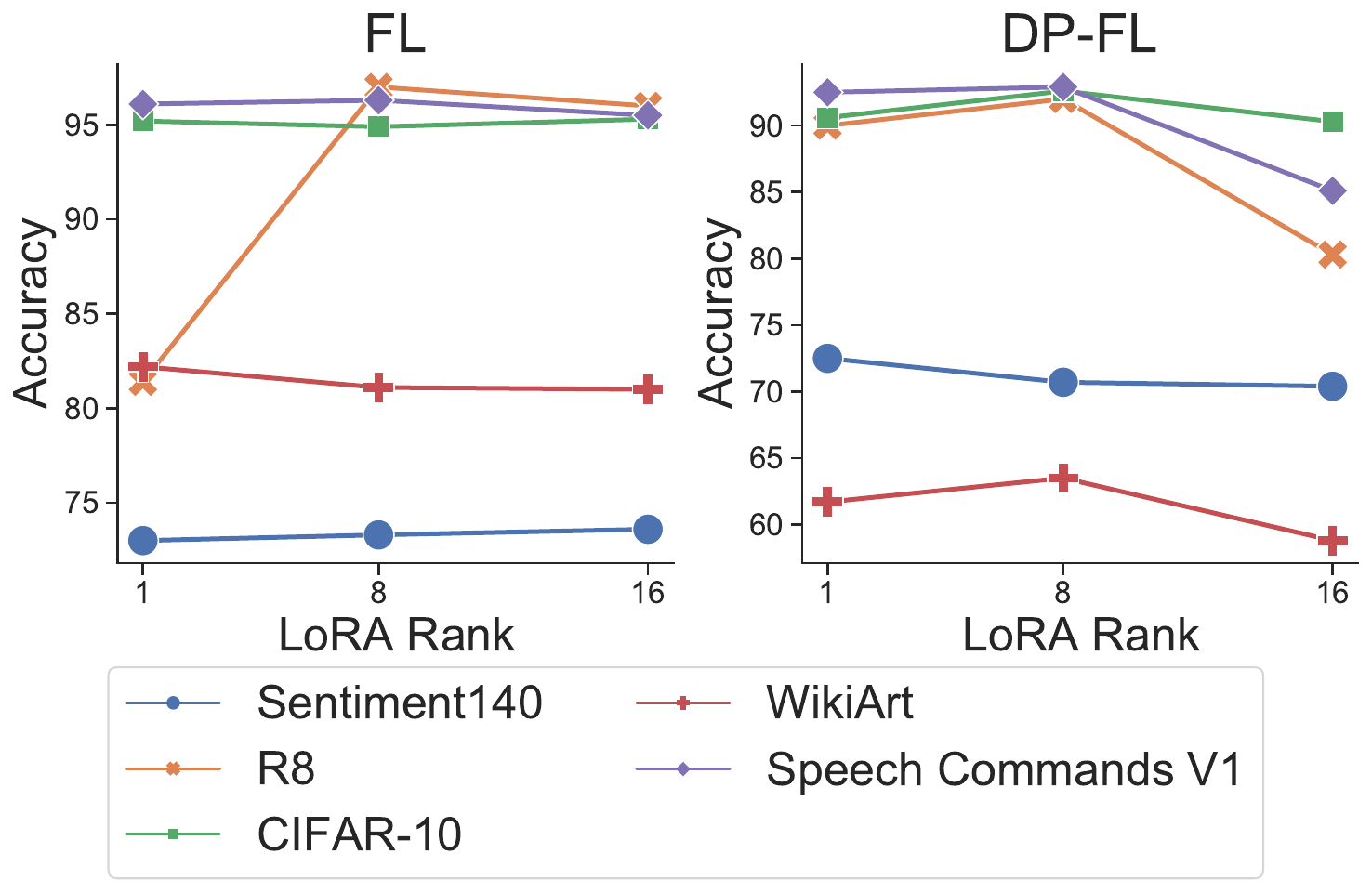}
    }
    \subfloat[\label{b}]{%
        \includegraphics[width=0.39\linewidth]{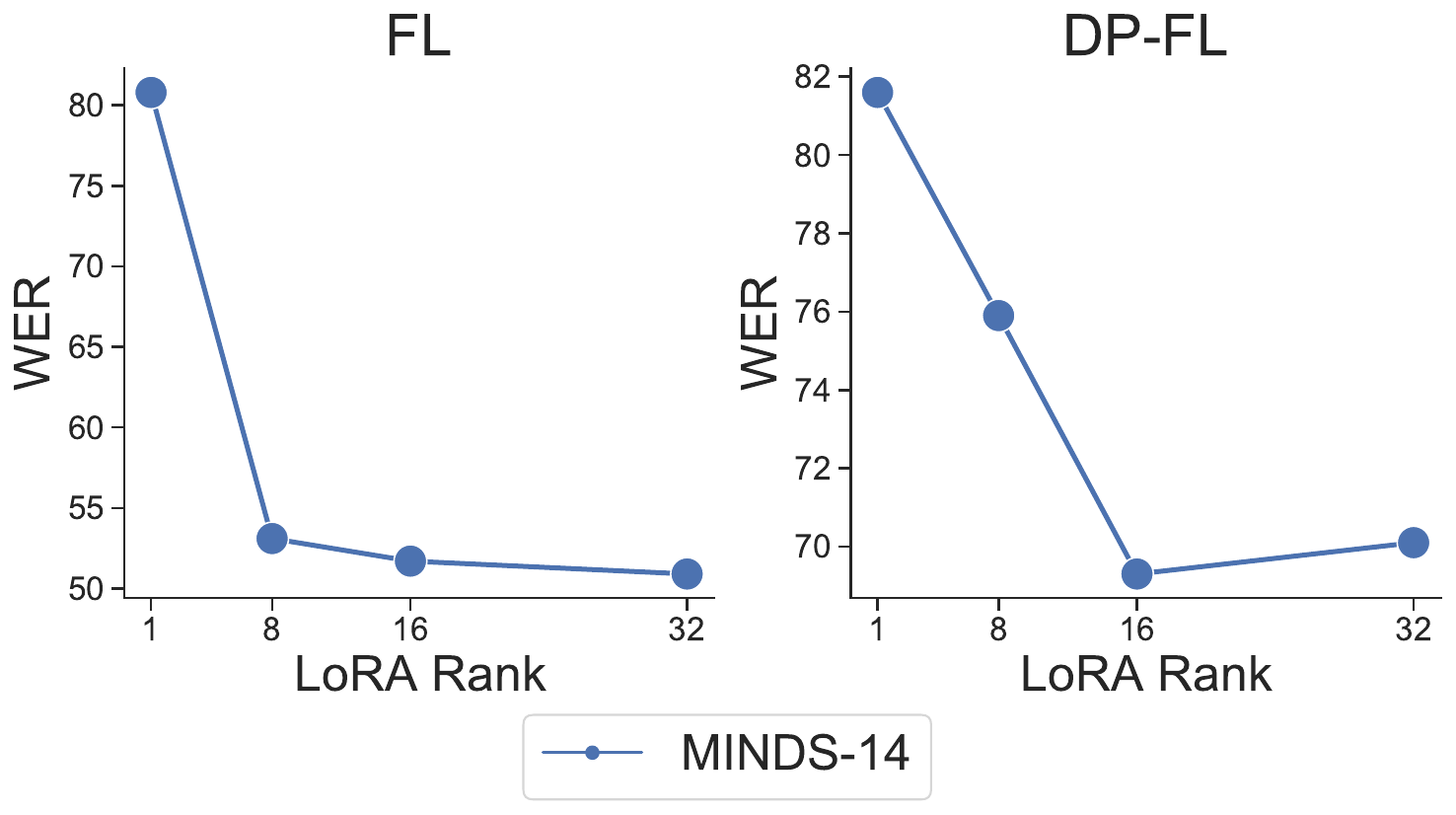}
    }
    \subfloat[\label{c}]{%
        \includegraphics[width=0.19\linewidth]{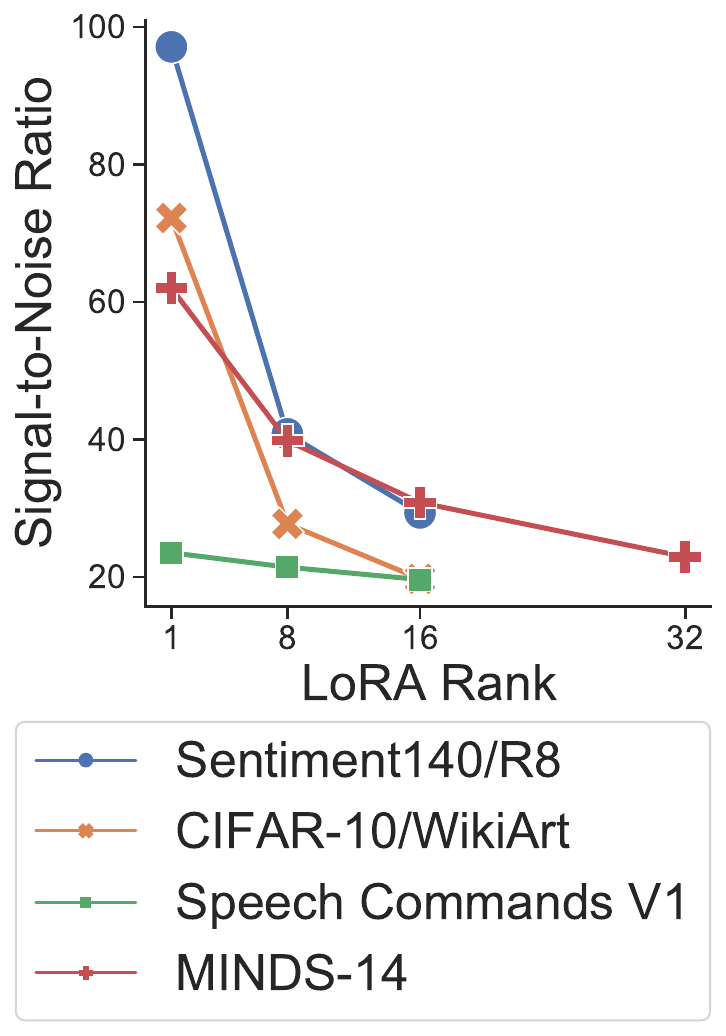}
    }
    \vspace*{-4pt}
    \caption{(a) and (b): LoRA performance on five different classification datasets under non-private and differentially private federated learning with increasing rank values; (c): Signal-to-noise ratio for LoRA with increasing rank values.}
    \label{fig_lora_rank}
    \vspace*{-10pt}
\end{figure*}

To better understand why DP-DyLoRA outperforms DP-LoRA, we plot the accuracy and WERs achieved by DP-LoRA with increasing rank values as well as the corresponding signal-to-noise ratio in Figure~\ref{fig_lora_rank}. As we can see, the best performance is achieved with $r$=8 in most cases. Although the model has the fewest number of trainable parameters with $r$=1 which subsequently leads to a higher signal-to-noise ratio as in Figure~\ref{fig_lora_rank}, the number of trainable parameters may also be insufficient for a given downstream task. This is especially the case when it comes to on-device models which are relatively small in size. On the other hand, with $r$=16 or $r$=32 as in the case of MINDS-14, the amount of added noise increases together with the number of trainable parameters which also hurts model performance. In other words, with our experimental settings, DP-LoRA with $r$=8 has a better balance between model expressiveness and the amount of added noise than other rank values. Regrading DP-DyLoRA, for $r_{min}$=1 and $r_{max}$=16, DP-DyLoRA has the same number of trainable parameters at the server side as DP-LoRA with $r$=16 and only updates a portion of the trainable weights. This makes it so that DP-DyLoRA has the same level of model expressiveness as DP-LoRA with $r$=16 while having similar signal-to-noise ratio as DP-LoRA with $r$=8. Hence, DP-DyLoRA achieves a better privacy-utility trade-off then DP-LoRA which leads to better DP-FL performance.

Another interesting finding from our results shown in Table~\ref{tab_peft_results_classification} and \ref{tab_peft_results_asr} is that DyLoRA actually performs slightly worse than LoRA in non-private FL. We notice that DyLoRA training tends to be unstable in FL setting with a reasonably large learning rate. This is especially obvious during the early training stage. These results therefore show that the partial update of LoRA weights makes DyLoRA more sensitive to data heterogeneity. One possible remedy is to apply gradient clipping, which is mandatory under DP-FL.

\noindent{\bf Take-aways}

\begin{list}{}{}
\item{DP-DyLoRA significantly outperforms existing DP-PEFT methods including DP-LoRA. Compared to LoRA or DyLoRA under non-private FL, DP-DyLoRA achieves less than 2\% accuracy drop for CV and NLU and less than 7\% WER increase for ASR with a strong privacy budget of $\bm{\epsilon}$=2 and 1 million clients.}
\end{list}

\vspace*{-8pt}

\section{Conclusion}
\label{sec_conclusion}

In this article, we present DP-DyLoRA, a novel differentially private federated learning (DP-FL) algorithm to mitigate the impact of noise addition under DP constraints. We show with empirical results that DP-DyLoRA outperforms the state-of-the-art method DP-LoRA on six datasets across three different domains with less than 2\% accuracy loss and 7\% word error rate (WER) increase from non-private LoRA (or DyLoRA) with a stringent privacy budget of $\epsilon=2$ and 1 million clients. In particular, our analysis shows that DP-DyLoRA suffers less from the trade-off between model expressiveness and amount of noise added due to DP guarantees which leads to better privacy-utility trade-offs under DP-FL.

\vspace*{-8pt}
\bibliographystyle{IEEEtran}
\bibliography{references}

\begin{thebibliography}{10}
\providecommand{\url}[1]{#1}
\csname url@samestyle\endcsname
\providecommand{\newblock}{\relax}
\providecommand{\bibinfo}[2]{#2}
\providecommand{\BIBentrySTDinterwordspacing}{\spaceskip=0pt\relax}
\providecommand{\BIBentryALTinterwordstretchfactor}{4}
\providecommand{\BIBentryALTinterwordspacing}{\spaceskip=\fontdimen2\font plus
\BIBentryALTinterwordstretchfactor\fontdimen3\font minus
  \fontdimen4\font\relax}
\providecommand{\BIBforeignlanguage}[2]{{%
\expandafter\ifx\csname l@#1\endcsname\relax
\typeout{** WARNING: IEEEtran.bst: No hyphenation pattern has been}%
\typeout{** loaded for the language `#1'. Using the pattern for}%
\typeout{** the default language instead.}%
\else
\language=\csname l@#1\endcsname
\fi
#2}}
\providecommand{\BIBdecl}{\relax}
\BIBdecl

\bibitem{transformer}
A.~Vaswani, N.~Shazeer, N.~Parmar, J.~Uszkoreit, L.~Jones, A.~N. Gomez, L.~u.
  Kaiser, and I.~Polosukhin, ``Attention is all you need,'' in \emph{Advances
  in Neural Information Processing Systems}, 2017.

\bibitem{mobilevit}
S.~Mehta and M.~Rastegari, ``Mobilevit: Light-weight, general-purpose, and
  mobile-friendly vision transformer,'' in \emph{International Conference on
  Learning Representations}, 2022.

\bibitem{mobile_device_transformer}
I.~Gim and J.~Ko, ``Memory-efficient dnn training on mobile devices.''\hskip
  1em plus 0.5em minus 0.4em\relax Association for Computing Machinery, 2022.

\bibitem{bert}
J.~Devlin, M.-W. Chang, K.~Lee, and K.~Toutanova, ``{BERT}: Pre-training of
  deep bidirectional transformers for language understanding,'' in
  \emph{Proceedings of the 2019 Conference of the North {A}merican Chapter of
  the Association for Computational Linguistics}, 2019.

\bibitem{gpt}
A.~Radford, K.~Narasimhan, T.~Salimans, I.~Sutskever \emph{et~al.}, ``Improving
  language understanding by generative pre-training,'' 2018.

\bibitem{federated_learning}
B.~McMahan, E.~Moore, D.~Ramage, S.~Hampson, and B.~A.~y. Arcas,
  ``Communication-efficient learning of deep networks from decentralized
  data,'' in \emph{Proceedings of the 20th International Conference on
  Artificial Intelligence and Statistics}, 2017.

\bibitem{dp_1}
C.~Dwork, F.~McSherry, K.~Nissim, and A.~Smith, ``Calibrating noise to
  sensitivity in private data analysis,'' in \emph{Theory of Cryptography},
  2006.

\bibitem{dp-sgd}
M.~Abadi, A.~Chu, I.~Goodfellow, H.~B. McMahan, I.~Mironov, K.~Talwar, and
  L.~Zhang, ``Deep learning with differential privacy,'' in \emph{Proceedings
  of the 2016 ACM SIGSAC Conference on Computer and Communications Security},
  2016.

\bibitem{privacy_amplification_via_subsampling_1}
B.~Balle, G.~Barthe, and M.~Gaboardi, ``Privacy amplification by subsampling:
  Tight analyses via couplings and divergences,'' in \emph{Advances in Neural
  Information Processing Systems}, 2018.

\bibitem{privacy_amplification_via_subsampling_2}
Y.-X. Wang, B.~Balle, and S.~P. Kasiviswanathan, ``Subsampled renyi
  differential privacy and analytical moments accountant,'' in \emph{The 22nd
  international conference on artificial intelligence and statistics}, 2019.

\bibitem{secure_aggregation_1}
K.~Bonawitz, V.~Ivanov, B.~Kreuter, A.~Marcedone, H.~B. McMahan, S.~Patel,
  D.~Ramage, A.~Segal, and K.~Seth, ``Practical secure aggregation for
  privacy-preserving machine learning,'' in \emph{ACM SIGSAC Conference on
  Computer and Communications Security}, 2017.

\bibitem{secure_aggregation_2}
J.~Bell, K.~A. Bonawitz, A.~Gascon, T.~Lepoint, and M.~Raykova, ``Secure
  single-server vector aggregation with (poly)logarithmic overhead,'' in
  \emph{Proceedings of the ACM SIGSAC Conference on Computer and Communications
  Security}, 2020.

\bibitem{secure_sum}
S.~Goryczka and L.~Xiong, ``A comprehensive comparison of multiparty secure
  additions with differential privacy,'' \emph{IEEE Transactions on Dependable
  and Secure Computing}, 2017.

\bibitem{noise_magnitude_model_size_1}
R.~Bassily, A.~Smith, and A.~Thakurta, ``Private empirical risk minimization:
  Efficient algorithms and tight error bounds,'' in \emph{2014 IEEE 55th Annual
  Symposium on Foundations of Computer Science}, 2014.

\bibitem{noise_magnitude_model_size_3}
M.~Abadi, A.~Chu, I.~Goodfellow, H.~B. McMahan, I.~Mironov, K.~Talwar, and
  L.~Zhang, ``Deep learning with differential privacy,'' in \emph{Proceedings
  of the 2016 ACM SIGSAC Conference on Computer and Communications Security},
  2016.

\bibitem{dp-fl-peft_1}
M.~Xu, C.~Song, Y.~Tian, N.~Agrawal, F.~Granqvist, R.~van Dalen, X.~Zhang,
  A.~Argueta, S.~Han, Y.~Deng, L.~Liu, A.~Walia, and A.~Jin, ``Training
  large-vocabulary neural language models by private federated learning for
  resource-constrained devices,'' in \emph{IEEE International Conference on
  Acoustics, Speech and Signal Processing (ICASSP)}, 2023.

\bibitem{dp-fl-peft_2}
H.~Zhao, W.~Du, F.~Li, P.~Li, and G.~Liu, ``Fedprompt: Communication-efficient
  and privacy-preserving prompt tuning in federated learning,'' in \emph{ICASSP
  2023-2023 IEEE International Conference on Acoustics, Speech and Signal
  Processing (ICASSP)}.\hskip 1em plus 0.5em minus 0.4em\relax IEEE, 2023, pp.
  1--5.

\bibitem{dp-fl-peft_3}
C.~Xie, D.-A. Huang, W.~Chu, D.~Xu, C.~Xiao, B.~Li, and A.~Anandkumar,
  ``Perada: Parameter-efficient and generalizable federated learning
  personalization with guarantees,'' \emph{arXiv:2302.06637}, 2023.

\bibitem{iot_millions_of_clients}
M.~Yun and B.~Yuxin, ``Research on the architecture and key technology of
  internet of things (iot) applied on smart grid,'' in \emph{2010 International
  Conference on Advances in Energy Engineering}, 2010, pp. 69--72.

\bibitem{adapter_1}
N.~Houlsby, A.~Giurgiu, S.~Jastrzebski, B.~Morrone, Q.~De~Laroussilhe,
  A.~Gesmundo, M.~Attariyan, and S.~Gelly, ``Parameter-efficient transfer
  learning for {NLP},'' in \emph{Proceedings of the 36th International
  Conference on Machine Learning}, 2019.

\bibitem{adapter_3}
J.~Pfeiffer, A.~R{\"u}ckl{\'e}, C.~Poth, A.~Kamath, I.~Vuli{\'c}, S.~Ruder,
  K.~Cho, and I.~Gurevych, ``{A}dapter{H}ub: A framework for adapting
  transformers,'' in \emph{Proceedings of the 2020 Conference on Empirical
  Methods in Natural Language Processing: System Demonstrations}, 2020.

\bibitem{compacter}
R.~K. mahabadi, J.~Henderson, and S.~Ruder, ``Compacter: Efficient low-rank
  hypercomplex adapter layers,'' in \emph{Advances in Neural Information
  Processing Systems}, 2021.

\bibitem{bitfit}
E.~Ben~Zaken, Y.~Goldberg, and S.~Ravfogel, ``{B}it{F}it: Simple
  parameter-efficient fine-tuning for transformer-based masked
  language-models,'' in \emph{Proceedings of the 60th Annual Meeting of the
  Association for Computational Linguistics (Volume 2: Short Papers)}, 2022.

\bibitem{lora}
E.~J. Hu, Y.~Shen, P.~Wallis, Z.~Allen-Zhu, Y.~Li, S.~Wang, L.~Wang, and
  W.~Chen, ``Lo{RA}: Low-rank adaptation of large language models,'' in
  \emph{International Conference on Learning Representations}, 2022.

\bibitem{dp-lora}
D.~Yu, S.~Naik, A.~Backurs, S.~Gopi, H.~A. Inan, G.~Kamath, J.~Kulkarni, Y.~T.
  Lee, A.~Manoel, L.~Wutschitz, S.~Yekhanin, and H.~Zhang, ``Differentially
  private fine-tuning of language models,'' in \emph{International Conference
  on Learning Representations}, 2022.

\bibitem{fl-peft_1}
J.~Chen, W.~Xu, S.~Guo, J.~Wang, J.~Zhang, and H.~Wang, ``Fedtune: A deep dive
  into efficient federated fine-tuning with pre-trained transformers,''
  \emph{arXiv preprint arXiv:2211.08025}, 2022.

\bibitem{dylora}
M.~Valipour, M.~Rezagholizadeh, I.~Kobyzev, and A.~Ghodsi, ``{D}y{L}o{RA}:
  Parameter-efficient tuning of pre-trained models using dynamic search-free
  low-rank adaptation,'' in \emph{Conference of the European Chapter of the
  Association for Computational Linguistics}, 2023.

\bibitem{loha}
N.~Hyeon-Woo, M.~Ye-Bin, and T.-H. Oh, ``Fedpara: Low-rank hadamard product for
  communication-efficient federated learning,'' in \emph{International
  Conference on Learning Representations}, 2022.

\bibitem{adalora}
Q.~Zhang, M.~Chen, A.~Bukharin, P.~He, Y.~Cheng, W.~Chen, and T.~Zhao,
  ``Adaptive budget allocation for parameter-efficient fine-tuning,'' in
  \emph{International Conference on Learning Representations}, 2023.

\bibitem{fedavg}
B.~McMahan, E.~Moore, D.~Ramage, S.~Hampson, and B.~A.~y. Arcas,
  ``{Communication-Efficient Learning of Deep Networks from Decentralized
  Data},'' in \emph{Proceedings of the 20th International Conference on
  Artificial Intelligence and Statistics}, 2017.

\bibitem{fedsgd}
J.~Chen*, X.~Pan*, R.~Monga, S.~Bengio, and R.~Jozefowicz, ``Revisiting
  distributed synchronous {SGD},'' in \emph{ICLR Workshop Track}, 2017.

\bibitem{scaffold}
S.~P. Karimireddy, S.~Kale, M.~Mohri, S.~Reddi, S.~Stich, and A.~T. Suresh,
  ``Scaffold: Stochastic controlled averaging for federated learning,'' in
  \emph{International conference on machine learning}, 2020.

\bibitem{q-ffl}
T.~Li, M.~Sanjabi, A.~Beirami, and V.~Smith, ``Fair resource allocation in
  federated learning,'' in \emph{International Conference on Learning
  Representations}, 2020.

\bibitem{fedopt}
S.~J. Reddi, Z.~Charles, M.~Zaheer, Z.~Garrett, K.~Rush, J.~Kone{\v{c}}n{\'y},
  S.~Kumar, and H.~B. McMahan, ``Adaptive federated optimization,'' in
  \emph{International Conference on Learning Representations}, 2021.

\bibitem{fedprox}
T.~Li, A.~K. Sahu, M.~Zaheer, M.~Sanjabi, A.~Talwalkar, and V.~Smith,
  ``Federated optimization in heterogeneous networks,'' \emph{Proceedings of
  Machine learning and systems}, vol.~2, pp. 429--450, 2020.

\bibitem{fl_niid_1}
Y.~Zhao, M.~Li, L.~Lai, N.~Suda, D.~Civin, and V.~Chandra, ``Federated learning
  with non-iid data,'' \emph{arXiv preprint arXiv:1806.00582}, 2018.

\bibitem{fl_niid_2}
X.~Li, K.~Huang, W.~Yang, S.~Wang, and Z.~Zhang, ``On the convergence of fedavg
  on non-iid data,'' in \emph{International Conference on Learning
  Representations}, 2020.

\bibitem{fl_niid_setting}
J.~Kone{\v c}n{\'y}, H.~McMahan, D.~Ramage, and P.~Richt{\'a}rik, ``Federated
  optimization: Distributed machine learning for on-device intelligence,''
  ArXiv, Tech. Rep., 2016.

\bibitem{adversary_1}
M.~Nasr, R.~Shokri, and A.~Houmansadr, ``Comprehensive privacy analysis of deep
  learning: Passive and active white-box inference attacks against centralized
  and federated learning,'' in \emph{2019 IEEE symposium on security and
  privacy (SP)}.\hskip 1em plus 0.5em minus 0.4em\relax IEEE, 2019, pp.
  739--753.

\bibitem{adversary_2}
L.~Zhu, Z.~Liu, and S.~Han, ``Deep leakage from gradients,'' in \emph{Advances
  in Neural Information Processing Systems}, 2019.

\bibitem{adversary_3}
J.~Geiping, H.~Bauermeister, H.~Dr\"{o}ge, and M.~Moeller, ``Inverting
  gradients - how easy is it to break privacy in federated learning?'' in
  \emph{Advances in Neural Information Processing Systems}, 2020.

\bibitem{dp-fedavg}
H.~B. McMahan, D.~Ramage, K.~Talwar, and L.~Zhang, ``Learning differentially
  private recurrent language models,'' in \emph{International Conference on
  Learning Representations}, 2018.

\bibitem{moments_accountant}
M.~Abadi, A.~Chu, I.~Goodfellow, H.~B. McMahan, I.~Mironov, K.~Talwar, and
  L.~Zhang, ``Deep learning with differential privacy,'' in \emph{Proceedings
  of the 2016 ACM SIGSAC Conference on Computer and Communications Security},
  2016.

\bibitem{gaussian_mechanism}
C.~Dwork, A.~Roth \emph{et~al.}, ``The algorithmic foundations of differential
  privacy,'' \emph{Foundations and Trends{\textregistered} in Theoretical
  Computer Science}, vol.~9, no. 3--4, pp. 211--407, 2014.

\bibitem{ghost_clipping}
X.~Li, F.~Tramer, P.~Liang, and T.~Hashimoto, ``Large language models can be
  strong differentially private learners,'' in \emph{International Conference
  on Learning Representations}, 2022.

\bibitem{peft_sequential_parallel}
J.~He, C.~Zhou, X.~Ma, T.~Berg-Kirkpatrick, and G.~Neubig, ``Towards a unified
  view of parameter-efficient transfer learning,'' in \emph{International
  Conference on Learning Representations}, 2022.

\bibitem{dp_2}
C.~Dwork, ``Differential privacy,'' in \emph{International colloquium on
  automata, languages, and programming}.\hskip 1em plus 0.5em minus 0.4em\relax
  Springer, 2006, pp. 1--12.

\bibitem{dp_3}
C.~Dwork, A.~Roth \emph{et~al.}, ``The algorithmic foundations of differential
  privacy,'' \emph{Foundations and Trends{\textregistered} in Theoretical
  Computer Science}, vol.~9, no. 3--4, pp. 211--407, 2014.

\bibitem{secure_aggregation_exact_sum}
P.~Kairouz, Z.~Liu, and T.~Steinke, ``The distributed discrete gaussian
  mechanism for federated learning with secure aggregation,'' in
  \emph{Proceedings of the 38th International Conference on Machine Learning},
  2021.

\bibitem{rdp}
I.~Mironov, ``R{\'e}nyi differential privacy,'' in \emph{2017 IEEE 30th
  computer security foundations symposium (CSF)}.\hskip 1em plus 0.5em minus
  0.4em\relax IEEE, 2017, pp. 263--275.

\bibitem{rdp_accountant}
B.~Balle, G.~Barthe, M.~Gaboardi, J.~Hsu, and T.~Sato, ``Hypothesis testing
  interpretations and {R}enyi differential privacy,'' in \emph{International
  Conference on Artificial Intelligence and Statistics}.\hskip 1em plus 0.5em
  minus 0.4em\relax PMLR, 2020.

\bibitem{laplace_mechanism}
C.~Dwork, F.~McSherry, K.~Nissim, and A.~Smith, ``Calibrating noise to
  sensitivity in private data analysis,'' in \emph{Theory of Cryptography:
  Third Theory of Cryptography Conference, TCC 2006, New York, NY, USA, March
  4-7, 2006. Proceedings 3}.\hskip 1em plus 0.5em minus 0.4em\relax Springer,
  2006, pp. 265--284.

\bibitem{flair}
C.~Song, F.~Granqvist, and K.~Talwar, ``{FLAIR}: Federated learning annotated
  image repository,'' in \emph{Thirty-sixth Conference on Neural Information
  Processing Systems Datasets and Benchmarks Track}, 2022.

\bibitem{sent140_preprocessing}
R.~H{\"o}nig, Y.~Zhao, and R.~Mullins, ``{DA}da{Q}uant: Doubly-adaptive
  quantization for communication-efficient federated learning,'' in
  \emph{Proceedings of the 39th International Conference on Machine Learning},
  2022.

\bibitem{reuters-21578}
D.~Lewis, ``{Reuters-21578 Text Categorization Collection},'' UCI Machine
  Learning Repository, 1997, {DOI}: https://doi.org/10.24432/C52G6M.

\bibitem{cifar-10}
A.~Krizhevsky, G.~Hinton \emph{et~al.}, ``Learning multiple layers of features
  from tiny images,'' 2009.

\bibitem{wikiart}
B.~Saleh and A.~Elgammal, ``Large-scale classification of fine-art paintings:
  Learning the right metric on the right feature,'' \emph{arXiv preprint
  arXiv:1505.00855}, 2015.

\bibitem{tiny_images}
A.~Torralba, R.~Fergus, and W.~T. Freeman, ``80 million tiny images: A large
  data set for nonparametric object and scene recognition,'' \emph{IEEE
  Transactions on Pattern Analysis and Machine Intelligence}, vol.~30, no.~11,
  pp. 1958--1970, 2008.

\bibitem{fl_llm_size}
J.~H. Ro, T.~Breiner, L.~McConnaughey, M.~Chen, A.~T. Suresh, S.~Kumar, and
  R.~Mathews, ``Scaling language model size in cross-device federated
  learning,'' in \emph{ACL 2022 Workshop on Federated Learning for Natural
  Language Processing}, 2022.

\bibitem{fl_transformer_3}
X.~Zhang, B.~Song, M.~Honarkhah, J.~Ding, and M.~Hong, ``Building large machine
  learning models from small distributed models: A layer matching approach,''
  in \emph{Workshop on Federated Learning: Recent Advances and New Challenges
  (in Conjunction with NeurIPS 2022)}, 2022.

\bibitem{dp-scaffold}
M.~Noble, A.~Bellet, and A.~Dieuleveut, ``Differentially private federated
  learning on heterogeneous data,'' in \emph{The 25th international conference
  on artificial intelligence and statistics}, 2022.

\bibitem{fl-dp-ftrl}
Z.~Xu, Y.~Zhang, G.~Andrew, C.~Choquette, P.~Kairouz, B.~Mcmahan, and
  J.~Rosenstock, ``Federated learning of gboard language models with
  differential privacy,'' in \emph{Proceedings of the 61st Annual Meeting of
  the Association for Computational Linguistics (Volume 5: Industry Track)},
  2023, pp. 629--639.

\bibitem{vit}
A.~Dosovitskiy, L.~Beyer, A.~Kolesnikov, D.~Weissenborn, X.~Zhai,
  T.~Unterthiner, M.~Dehghani, M.~Minderer, G.~Heigold, S.~Gelly, J.~Uszkoreit,
  and N.~Houlsby, ``An image is worth 16x16 words: Transformers for image
  recognition at scale,'' in \emph{International Conference on Learning
  Representations}, 2021.

\bibitem{hubert}
W.-N. Hsu, B.~Bolte, Y.-H.~H. Tsai, K.~Lakhotia, R.~Salakhutdinov, and
  A.~Mohamed, ``Hubert: Self-supervised speech representation learning by
  masked prediction of hidden units,'' \emph{IEEE/ACM Trans. Audio, Speech and
  Lang. Proc.}, 2021.

\bibitem{distilhubert}
H.-J. Chang, S.-w. Yang, and H.-y. Lee, ``Distilhubert: Speech representation
  learning by layer-wise distillation of hidden-unit bert,'' in \emph{ICASSP
  2022 - 2022 IEEE International Conference on Acoustics, Speech and Signal
  Processing (ICASSP)}, 2022, pp. 7087--7091.

\bibitem{llama}
H.~Touvron, T.~Lavril, G.~Izacard, X.~Martinet, M.-A. Lachaux, T.~Lacroix,
  B.~Rozi{\`e}re, N.~Goyal, E.~Hambro, F.~Azhar \emph{et~al.}, ``Llama: Open
  and efficient foundation language models,'' \emph{arXiv preprint
  arXiv:2302.13971}, 2023.

\bibitem{llama_2}
H.~Touvron, L.~Martin, K.~Stone, P.~Albert, A.~Almahairi, Y.~Babaei,
  N.~Bashlykov, S.~Batra, P.~Bhargava, S.~Bhosale \emph{et~al.}, ``Llama 2:
  Open foundation and fine-tuned chat models,'' \emph{arXiv preprint
  arXiv:2307.09288}, 2023.

\bibitem{gpt-4}
OpenAI, ``Gpt-4 technical report,'' \emph{ArXiv}, 2023.

\bibitem{dirichlet_alpha_1}
T.~Lin, L.~Kong, S.~U. Stich, and M.~Jaggi, ``Ensemble distillation for robust
  model fusion in federated learning,'' 2020.

\bibitem{dirichlet_alpha_2}
M.~Luo, F.~Chen, D.~Hu, Y.~Zhang, J.~Liang, and J.~Feng, ``No fear of
  heterogeneity: Classifier calibration for federated learning with non-{IID}
  data,'' in \emph{Advances in Neural Information Processing Systems}, 2021.

\bibitem{dirichlet_alpha_3}
H.-Y. Chen and W.-L. Chao, ``On bridging generic and personalized federated
  learning for image classification,'' in \emph{ICLR}, 2022.

\bibitem{laplace_reason_why_exclude}
S.~Casacuberta, M.~Shoemate, S.~Vadhan, and C.~Wagaman, ``Widespread
  underestimation of sensitivity in differentially private libraries and how to
  fix it,'' in \emph{Proceedings of the 2022 ACM SIGSAC Conference on Computer
  and Communications Security}, 2022.

\bibitem{sampling_rate_1}
Y.~Yeganeh, A.~Farshad, N.~Navab, and S.~Albarqouni, ``Inverse distance
  aggregation for federated learning with non-iid data,'' in \emph{Domain
  Adaptation and Representation Transfer, and Distributed and Collaborative
  Learning: Second MICCAI Workshop, DART 2020, and First MICCAI Workshop, DCL
  2020, Held in Conjunction with MICCAI 2020, Lima, Peru, October 4--8, 2020,
  Proceedings 2}, 2020.

\bibitem{sampling_rate_2}
J.~Hernandez \emph{et~al.}, ``Privacy-first health research with federated
  learning,'' \emph{medRxiv}, 2020.

\end{thebibliography}
% \end{thebibliography}

\vfill

\end{document}